\documentclass{article} 
\usepackage{iclr2025_conference,times}
\iclrfinalcopy

\usepackage{amsmath,amsfonts,bm}

\def\eqref#1{equation~\ref{#1}}

\def\1{\bm{1}}

\DeclareMathAlphabet{\mathsfit}{\encodingdefault}{\sfdefault}{m}{sl}
\SetMathAlphabet{\mathsfit}{bold}{\encodingdefault}{\sfdefault}{bx}{n}

\usepackage{hyperref}
\usepackage{url}
\usepackage{booktabs}
\usepackage{hyperref}
\usepackage{graphicx}
\usepackage{subcaption} 
\usepackage{float}      
\usepackage{array}
\usepackage{fancyvrb}

\newcolumntype{C}[1]{>{\centering\arraybackslash}m{#1}}

\title{Llama Scope: Extracting Millions of Features from Llama-3.1-8B with Sparse Autoencoders}

\author{Zhengfu He\textsuperscript{1,2}\quad Wentao Shu\textsuperscript{1}\quad Xuyang Ge\textsuperscript{1} \\
\textbf{Lingjie Chen\textsuperscript{1}\quad Junxuan Wang\textsuperscript{1}\quad Yunhua Zhou\textsuperscript{1,3}\quad Frances Liu} \\
\textbf{Qipeng Guo\textsuperscript{1,2,3}\quad Xuanjing Huang\textsuperscript{1}\quad Zuxuan Wu\textsuperscript{1}\quad Yu-Gang Jiang\textsuperscript{1}\quad Xipeng Qiu\textsuperscript{1,2}} \\
\\
\textsuperscript{1}Fudan University \\
\textsuperscript{2}Shanghai Innovation Institute \\
\textsuperscript{3}Shanghai Artificial Intelligence Laboratory \\
\\
\texttt{\{zfhe19, xpqiu\}@fudan.edu.cn} \\
}

\begin{document}

\maketitle

\begin{abstract} 
   Sparse Autoencoders (SAEs) have emerged as a powerful unsupervised method for extracting sparse 
   representations from language models, yet scalable training remains a significant challenge. 
   We introduce a suite of 256 SAEs, trained on each layer and sublayer of the Llama-3.1-8B-Base model, 
   with 32K and 128K features. Modifications to a state-of-the-art SAE variant, Top-K SAEs, are evaluated 
   across multiple dimensions. In particular, we assess the generalizability of SAEs trained on base 
   models to longer contexts and fine-tuned models. Additionally, we analyze the geometry of learned SAE 
   latents, confirming that \emph{feature splitting} enables the discovery of new features. The Llama Scope 
   SAE checkpoints are publicly available at~\url{https://huggingface.co/fnlp/Llama-Scope}, alongside our 
   scalable training, interpretation, and visualization tools at
   \url{https://github.com/OpenMOSS/Language-Model-SAEs}. These contributions aim to advance the 
   open-source Sparse Autoencoder ecosystem and support mechanistic interpretability research by 
   reducing the need for redundant SAE training.

\end{abstract}

\section{Introduction}

Mechanistic interpretability has long grappled with the challenge of identifying interpretable primitives 
within language models. Despite this, researchers have demonstrated that the structure of network 
representations tends to be linear, sparse, and decomposable~\citep{mikolov2013w2v, olah2020zoom, 
elhage2022superposition, gurnee2023sparseprobing}. Ideally, the model's components, such as neurons 
and attention heads, would correspond directly to interpretable features of the input. However, due 
to superposition~\citep{arora2018superposition, elhage2022superposition} and the misalignment of linear 
features with neuron bases~\citep{Elhage2023privileged}, this is not typically the case.

Sparse Autoencoders (SAEs)~\citep{bricken2023monosemanticity, huben2024SAE, templeton2024scaling}
offer a promising approach to addressing superposition. The features extracted by SAEs exhibit high
monosemanticity and causal relevance, allowing them to capture significantly more features than
neuron-based methods in pretrained Transformer models\citep{olah2020zoom, Bills2023autointerp, huben2024SAE}.
This technique aids in identifying the latent variables within neural networks, providing anchor points
for reverse engineering. These features may also prove useful in addressing model hallucination and
mitigating safety-relevant behaviors.

Despite these advances, research in this area remains somewhat disconnected from industrial-scale 
language models. While recent efforts have begun training and investigating SAEs on models exceeding 
8 billion parameters~\citep{templeton2024scaling, engels2024notalllinear, gao2024oaisae, lieberum2024gemmascope}, 
rigorous interpretability research on such large models is still in its early stages. Comprehensive 
analyses often require SAEs trained across multiple sites (e.g., for SAE-based circuit 
analysis~\citep{he2024othellocircuits, marks2024sparsefeaturecircuits, ge2024hierattr}) or 
trained to handle multiple feature sizes (e.g., for feature splitting~\citep{bricken2023monosemanticity} 
and identifying ultra-rare features~\citep{templeton2024scaling}).

To address these challenges, we introduce Llama Scope, a suite of 256 Sparse Autoencoders trained on 
Llama-3.1-8B, a widely used open-source large language model.  Llama Scope aims to facilitate research in
mechanistic interpretability by providing ready-to-use, open-source SAE models, reducing the need 
for extensive retraining. We believe that this open-source ecosystem will serve as a \emph{common language} 
for researchers to communicate and share insights across the field.

The main contributions of this work are highlighted as follows:

\paragraph{Decomposing Every Activation Space of Llama-3.1-8B-Base.} We train SAEs on every
sublayer, namely post-MLP residual stream, attention output, MLP output, and Transcoders, for all 32
layers with 32K and 128K feature widths (Section~\ref{sec:training-position}). This results in 256
SAEs in total. Such a comprehensive suite of SAEs trained on industrial-scale language models opens
up new possibilities for both interpretability and LLM research. Following~\citet{lieberum2024gemmascope},
we list a number of exciting research directions that can be pursued with this suite in 
Section~\ref{sec:related-work}.

\paragraph{Improved TopK SAEs.} We make several modifications to the Top-K SAEs, including
incorporating the 2-norm of the decoder columns into the TopK computation, post-processing TopK SAEs
to JumpReLU variants, and introducing a K-annealing training schedule (Section~\ref{sec:improved_topk_sae}).

\paragraph{Comprehensive Evaluation.} Llama Scope SAEs are evaluated using a range of metrics, including
canonical metrics like sparsity-fidelity trade-off, latent firing frequency, and feature interpretability.
We also assess the generalizability of SAEs trained on base models to longer contexts and fine-tuned models
and analyze the geometry of learned SAE latents (Section~\ref{sec:evaluation}).

\paragraph{Disk-IO Friendly Training Infrastructure.} A mixed parallelism approach is employed to train
SAEs with a large number of features, which significantly reduces the memory bottleneck of SAE training
(Appendix~\ref{sec:infrastructure}).

\section{Conceptual and Technical Background}

\subsection{Attacking Superposition with Sparse Autoencoders}

The Sparse Autoencoder approach is motivated by the superposition hypothesis
\citep{arora2018superposition,elhage2022superposition}, which posits that
the representations learned by neural networks are composed of independent features with
the following properties:
\begin{itemize} 
   \item \textbf{Decomposability}: High-dimensional representations can be expressed as a combination of independent, interpretable features. 
   \item \textbf{Linearity}: Features are represented linearly, meaning the direction of each feature indicates the presence of a specific concept, while its magnitude reflects the concept’s strength. 
   \item \textbf{Sparsity}: The decomposition is sparse, with only a few features active at any given time. 
   \item \textbf{Overcompleteness}: The number of underlying features exceeds the dimensionality of the representation. 
\end{itemize}

This hypothesis combines several well-established concepts from the literature. 
For instance, in~\citet{olshausen96sparse}, it has been
shown that images can be sparsely represented by an overcomplete basis set of Gabor functions.
Similarly, \citet{mikolov2013w2v} showed that word embeddings exhibit a degree of
linear structure, exemplified by relationships such as $v_{\text{king}} - v_{\text{queen}} \approx
v_{\text{man}} - v_{\text{woman}}$. Linear probing~\citep{alain17probing, gurnee2023sparseprobing} 
also reflects this hypothesis, as it assumes the linearity of features in neural networks.

Despite some modifications~\citep{engels2024notalllinear} and counterexamples
~\citep{csordas2024counterLRH}, the superposition hypothesis remains a valuable framework
of how individual features are represented in neural networks. To overcome the superposition
problem, the field of mechanistic interpretability has been actively developing methods to
extract sparse and linear features from neural networks~\citep{elhage2022solu,bricken2023monosemanticity}, with 
Sparse Autoencoders being one of the most prominent approaches.

\subsection{Vanilla Sparse Autoencoders}\label{sec:sae}

Sparse Autoencoders are designed to reverse the effects of superposition by 
extracting features that are sparse, linear, and decomposable.

A Sparse Autoencoder typically consists of a single hidden layer. The input
$\textbf{x}$ is linearly mapped to a hidden layer $f(\textbf{x})$ followed by a non-linear
activation function.

\begin{equation}
   f(\textbf{x}) = \text{ReLU}(W^{enc}\textbf{x} + b^{enc}),
\end{equation}

Where $W^{enc} \in \mathbb{R}^{F \times D}$ is the weight matrix and $b^{enc} \in \mathbb{R}^F$ 
is the bias vector. The hidden layer is then linearly mapped back to the input space to 
reconstruct the input $\hat{\textbf{x}}$ with $W^{dec} \in \mathbb{R}^{D \times F}$
and a bias term $b^{dec} \in \mathbb{R}^D$:

\begin{equation}\label{eq:vanilla_decoder}
   \hat{\textbf{x}} = W^{dec}f(\textbf{x}) + b^{dec}.
\end{equation}

To ensure that the extracted features are sparse, a sparsity constraint is applied to the hidden layer. 
A common approach is L1 regularization, which encourages fewer active features while minimizing 
reconstruction error:

\begin{equation}
   \mathcal{L}=\mathcal{L}_\text{MSE} + \mathcal{L}_\text{Sparsity} = \lVert x-\hat{x}\rVert_2 + \lambda \sum_{i=1}^{F} \lVert f_i(\textbf{x})\rVert_1,
\end{equation}

Conceptually, the encoder decides which features are activated and to what degree, while 
the decoder maps these features back to the input space. The decoder columns
$W^{dec}_{:,i}$ form an overcomplete basis for the latent space, where each 
column's direction corresponds to a specific feature.

There are two primary ways to interpret the features extracted by SAEs. The \emph{encoder view} 
focuses on feature activations, identifying which inputs activate particular features and to what
extent. The \emph{decoder view}, by contrast, interprets the basis formed by the decoder columns, 
revealing the geometric structure of the latent space. This perspective provides insight into how 
neural networks encode complex information in a compressed form.

\subsection{Top-K Sparse Autoencoders}

We follow the approach of~\citet{gao2024oaisae}, which introduced Top-K Sparse Autoencoders (SAEs) 
as an improvement over vanilla SAEs (Section~\ref{sec:pareto}), demonstrating better performance 
on canonical metrics such as the sparsity-fidelity trade-off~\citep{gao2024oaisae}, as well as 
on novel evaluation metrics proposed by~\citet{anthropic24aug}.

Top-K SAEs enforce sparsity by selecting only the K most active features, $f_i(\textbf{x})$, 
for reconstruction, setting the remaining features to zero. The hidden layer of the SAE is 
thus defined as:

\begin{equation}\label{eq:encoder-oai-topk}
   f_i(\textbf{x}) = \text{TopK}(\text{ReLU}(W^{enc}_{i,:}\textbf{x} + b^{enc}_i)), i\in\{1, 2, \ldots, F\},
\end{equation}

The units in the hidden layer, $f_i(\textbf{x})$, are referred to as \emph{features}, 
representing the magnitude of the corresponding latent interpretability primitives.

As in the vanilla SAE, the hidden layer is then mapped back to the input space using a 
decoder matrix as in Equation~\ref{eq:vanilla_decoder}.
In addition, the decoder columns $W^{dec}_{:,i}$ are set to unit 2-norm after each minibatch 
to prevent the network from shrinking feature activations and compensating by increasing the 
norm of the corresponding decoder columns.

The training objective is to minimize the reconstruction error, which is simply the mean squared error:
$\mathcal{L} = \mathcal{L}_\text{MSE}=\lVert x-\hat{x}\rVert_2$.

\subsection{Modifications to Top-K SAEs}\label{sec:improved_topk_sae}

We introduce several modifications to the Top-K SAEs. Notably, we do not apply any 
auxiliary loss to \emph{revive} dead features, as this is rarely an issue in our SAEs.

\paragraph{Incorporating the 2-norm of Decoder Columns in Top-K Computation.}

One key modification involves incorporating the 2-norm of the decoder columns into the Top-K computation
to avoid normalizing decoder columns $W^{dec}_{:,i}$ to unit 2-norm after each minibatch, which adopts
the same motivation as~\citet{templeton2024scaling}.
Specifically, we redefine the activation of a feature, $f_i(\textbf{x})$, as the product of the feature's 
activation and the 2-norm of the corresponding decoder column, $\lVert W^{dec}_{:,i} \rVert_2$. The hidden 
layer is thus computed as:

\begin{align}\label{eq:encoder-our-topk}
   \begin{split}
      \text{Mask} &= \text{TopK}\left(\text{ReLU}(W^{enc}\textbf{x} + b^{enc})\cdot\begin{bmatrix} \|W^{dec}_{:,1}\|_2 \\ \|W^{dec}_{:,2}\|_2 \\ \vdots \\ \|W^{dec}_{:,F}\|_2 \end{bmatrix}\right),\\
      f_i(\textbf{x}) &= \text{Mask}_i \cdot (\text{ReLU}(W^{enc}_{i,:}\textbf{x} + b^{enc}_i)), i\in\{1, 2, \ldots, F\}.
   \end{split}
\end{align}

This modification also avoids the need to prune gradients parallel to the decoder columns 
in the Adam optimizer, which was previously required to prevent suboptimal training 
outcomes~\citep{bricken2023monosemanticity}. We conjecture that pruning gradients in this manner 
introduces noise into the Adam momentum, leading to less efficient SAE training. By incorporating 
the 2-norm into the Top-K computation, we achieve the same sparsity enforcement without the need 
for additional pruning.

It is important to note that during training, the 2-norm of the decoder columns is not fixed to 1,
which can conflict with its interpretation as the direction of the feature 
(Section~\ref{sec:sae}). We address this issue at inference time, as discussed in Section~\ref{sec:post-training}.

\paragraph{Post Processing TopK SAEs to JumpReLU Variants.}

Another modification is applied post-training. We introduce a threshold, $\theta$, to ensure that, 
on average, K features are activated across the training distribution, rather than exactly K 
features being activated for every input~\citep{anthropic24aug}. Features with activations exceeding 
this threshold retain their values, while those below are set to zero. This approach actually mirrors the 
JumpReLU activation function~\citep{erichson20jumprelu,rajamanoharan2024jumprelu} with 
its threshold set to $\theta$. Further details can be found in Section~\ref{sec:post-training}.

This variant combines the advantages of both Top-K and JumpReLU activations. Training with a 
predetermined L0 is more intuitive, as it allows researchers to set a desired L0 and observe 
the reconstruction loss during convergence. Additionally, thresholding during inference 
decouples the activation of features, meaning each feature is evaluated independently. 
This avoids scenarios where a feature remains inactive merely 
because other features are more strongly activated.

\paragraph{K-Annealing Training Schedule.}
In the early stages of training, we find it beneficial to gradually reduce the number of activated 
features from $D$ to $K$ over the first 10\% of training steps. This K-annealing schedule improves 
convergence by allowing the model to adjust more smoothly to sparse activations. We refer readers 
to Section~\ref{sec:optimization} for details.

\section{Training Llama Scope SAEs}

\begin{table}[h]
   \centering
   \resizebox{\textwidth}{!}{%
      \begin{tabular}{@{}C{3cm}C{3cm}C{3cm}C{3cm}C{3cm}@{}}
            \toprule
            & \textbf{Llama Scope} & \textbf{Scaling Monosemanticity} & \textbf{GPT-4 SAE} & \textbf{Gemma Scope} \\ 
            \midrule
            \textbf{Models} & \hspace{0.3cm}Llama-3.1 8B\newline(Open Source) & Claude-3.0 Sonnet (Proprietary) & \hspace{0.5cm}GPT-4\newline(Proprietary) & Gemma-2 2B \& 9B (Open Source) \\ 
            \midrule
            \textbf{SAE Training Data} & {SlimPajama} & Proprietary & Proprietary & Proprietary, Sampled from~\citet{mesnard2024gemma} \\ 
            \midrule
            \textbf{SAE Position (Layer)} & Every Layer & The Middle Layer & $\frac{5}{6}$ Late Layer & Every Layer \\ 
            \midrule
            \textbf{\hspace{0.3cm}SAE Position\newline(Site)} & R, A, M, TC & R & R & R, A, M, TC \\ 
            \midrule
            \textbf{\hspace{0.5cm}SAE Width\newline(\# Features)} & 32K, 128K & 1M, 4M, 34M & 128K, 1M, 16M & 16K, 64K, 128K, 256K - 1M (Partial) \\ 
            \midrule
            \textbf{\hspace{0.5cm}SAE Width\newline(Expansion Factor)} & 8x, 32x & Proprietary & Proprietary & 4.6x, 7.1x, 28.5x, 36.6x \\ 
            \midrule
            \textbf{Activation Function} & TopK-ReLU & ReLU & TopK-ReLU & JumpReLU \\ 
            \bottomrule
      \end{tabular}
   }
   \caption{An overview of existing work training SAEs on Language Models with more than 8B parameters.}\label{tab:comparison}
\end{table}

Table~\ref{tab:comparison} provides a broad overview of Llama Scope SAE suite, along with a 
comparison to recent work on training Sparse Autoencoders on models with more than 8 billion 
parameters~\citep{templeton2024scaling,gao2024oaisae,lieberum2024gemmascope}.

\subsection{Collecting Activations}\label{sec:collect_activations}
Llama Scope SAEs are trained on activations generated from Llama-3.1-8B using text data sampled 
from SlimPajama~\citep{cerebras2023slimpajama} with the proportions of each subset 
(e.g., Commoncrawl, C4, Github, Books, Arxiv, StackExchange) preserved. All activations and 
computations are performed using bfloat16 precision to optimize memory usage without sacrificing 
accuracy, as also demonstrated in~\citet{lieberum2024gemmascope} where bfloat16 and float32
activations are shown to have similar performance.

Each document is truncated to 1024 tokens, and a \textless beginoftext\textgreater~token
is prepended to the start of every document.
During training, the activations of the
\textless beginoftext\textgreater~and~\textless endoftext\textgreater~tokens are excluded from use.

It has been shown that the activations of different sequence positions of the same document
tend to be highly correlated~\citep{bricken2023monosemanticity}, potentially reducing the 
diversity of the training data. 
To address this, randomness is introduced by employing a buffer shuffling strategy: the 
buffer is refilled and shuffled at regular intervals (Appendix~\ref{sec:activation-buffer}).

\subsection{Training Position}\label{sec:training-position}

We train SAEs at each layer and sublayer of Llama-3.1-8B. Specifically, there are four 
potential training positions:

\begin{figure}[h]
   \begin{minipage}{0.5\textwidth}
       \centering
       \includegraphics[width=0.99\linewidth]{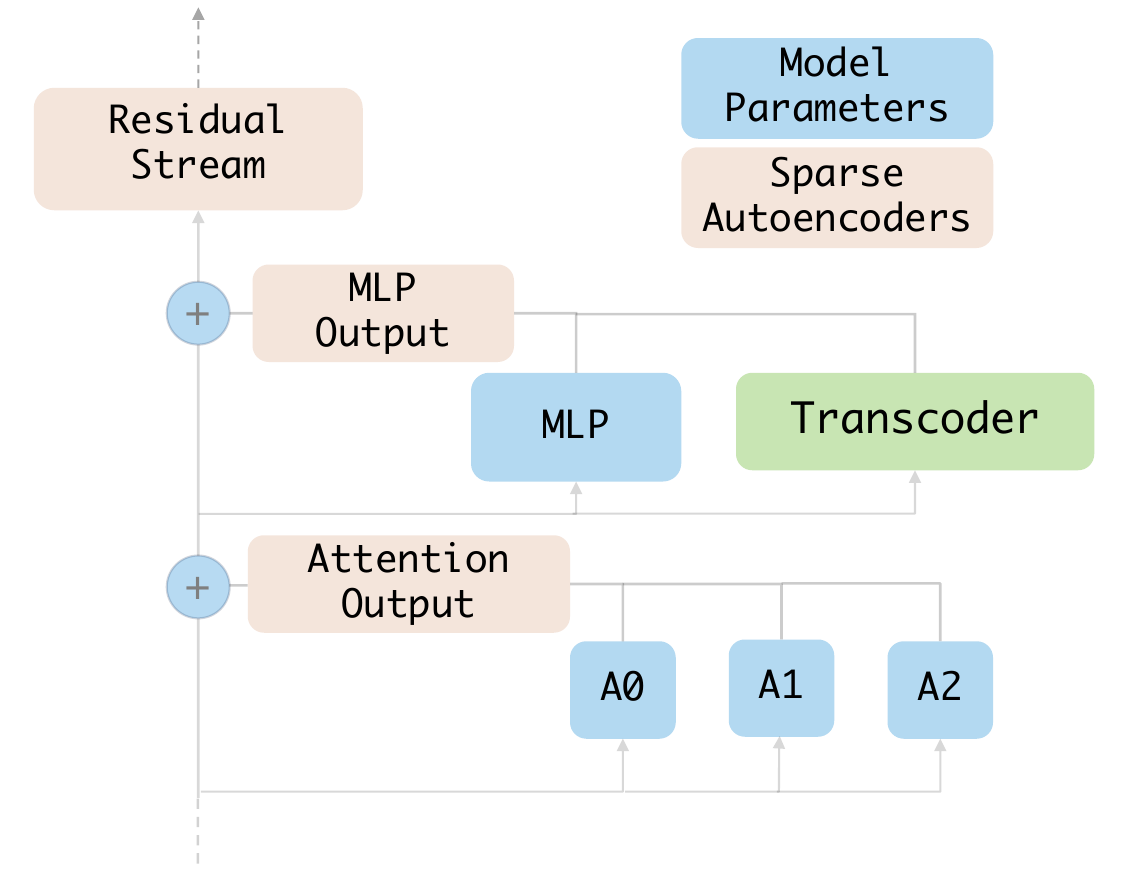}
       \caption{Four potential training positions in one Transformer Block.}\label{fig:training-position}
   \end{minipage}
   \begin{minipage}{0.5\textwidth}
      \begin{itemize} 
         \item \textbf{Post-MLP Residual Stream (R):} The residual stream after each Transformer block, which is the
         overall result of computations before this layer.
         \item \textbf{Attention Output (A):} The output of each attention layer.
         \item \textbf{MLP Output (M):} The output of each MLP layer. 
         \item \textbf{Transcoder (TC):} A sparse approximation of the MLP block designed for circuit 
         analysis~\citep{ge2024hierattr, dunefsky2024transcoders}. It differ from SAEs in that it
         takes in the layer-normalized residual stream as input and attempts to predict MLP output.
      \end{itemize}
   \end{minipage}
\end{figure}

Llama-3.1-8B consists of 32 layers, resulting in 128 possible training positions when considering all four options.
For each training position, SAEs are trained with 8x (32K features) and 32x (128K features) the 
dimension of the hidden size of Llama-3.1-8B.

Throughout this paper, we adopt the naming convention introduced by~\citet{he2024othellocircuits} to label the SAEs 
as~\texttt{L[Layer][Position]-[Expansion]x-[TopK,Vanilla]}.
For example, a Top-K SAE trained on the post-MLP residual stream of layer 1 in Llama-3.1-8B, 
with an 8x expansion of the hidden size is named as~\texttt{L1R-8x-TopK}.

\subsection{Training Details}

\subsubsection{Initialization}

The decoder columns $W^{dec}_{:,i}$ are initialized with Kaiming uniform~\citep{he2016resnet} 
and normalized to have 2-norm of $\sqrt{\frac{2D}{F}} = \sqrt{\frac{2}{\text{expansion factor}}}$. 
The encoder weights $W^{enc}$ are initially set to the transpose of the decoder weights $W^{dec}$, 
though their parameters are untied after initialization. Both the encoder bias $b^{enc}$ and decoder 
bias $b^{dec}$ are initialized to zero.

This initialization strategy allows the SAE to start with a near-zero reconstruction loss, a widely 
observed benefit in SAE training~\citep{templeton2024scaling, gao2024oaisae, lieberum2024gemmascope}.

For transcoders, the encoder weights are not initialized as the transpose of the decoder weights, 
as the input and output follow different distributions.

\subsubsection{Optimization}\label{sec:optimization}
We train the SAEs using the Adam optimizer with $\beta_1 = 0.9$, $\beta_2 = 0.999$, and
$\epsilon = 10^{-8}$. The learning rate is set to 8e-4 for all SAEs, with a warm-up from 
0 to the target rate over the first 10K steps, followed by a linear decay to 0 during 
the final 20\% of training steps\footnote{We find that learning rate 
decay significantly benefits SAE training.}.

We also observed that fixing the number of activated features to K throughout training 
can result in some features not being activated until halfway of the training process. 
To address this, we gradually reduce the number of activated features from $D$ to $K$ 
during the first 10\% of training steps. This approach, similar to L1 coefficient 
warmup in L1 regularization~\citep{templeton2024scaling}, ensures more even feature 
activation throughout the training.

\subsubsection{Input and Output Normalization}
The 2-norm of the input $\lVert \textbf{x} \rVert_2$ varies across different positions.
To ensure consistent hyperparameters, we normalize the input $\textbf{x}$ to have a 2-norm of
$\sqrt{D}$ before passing it into the SAE, and the reconstruction loss is computed with 
the same normalization.
For Transcoders, we normalize both the input and output to have a 2-norm of $\sqrt{D}$.

Formally, the input normalization is performed as follows:
\begin{align*}
   \textbf{x}_\text{in} &\leftarrow \textbf{x}_\text{in} \cdot S_{in},\\
   \textbf{x}_\text{out} &\leftarrow \textbf{x}_\text{out} \cdot S_{out},
\end{align*}

where $S_{in}$ and $S_{out}$ are the scalar normalization factors defined as
$S_{in} = \sqrt{D} / \mathbb{E} (\lvert \lvert \textbf{x}_\text{in} \rvert \rvert_2)$ and $S_{out} = 
\sqrt{D} / \mathbb{E} (\lvert \lvert \textbf{x}_\text{out} \rvert \rvert_2)$. These two factors 
differ only when training transcoders.

\subsubsection{Post-Training Processing}\label{sec:post-training}

After training, we rescale the learned SAE weights to account for the input and output normalization 
that was applied during training. This is done using the following transformations:

\begin{align*}
   W^{dec}_{:,i} &\leftarrow \frac{W^{dec}_{:,i}}{S_{out}} \cdot S_{in},\\
   b^{dec} &\leftarrow \frac{b^{dec}}{S_{out}},\\
   b^{enc} &\leftarrow \frac{b^{enc}}{S_{in}}.
\end{align*}

For standard SAEs, where $S_{in} = S_{out}$, the decoder weights remain unchanged.

To simplify analysis, we further adjust the decoder columns $W^{dec}_{:,i}$ to have unit 2-norm. 
This ensures that each column indicates only the direction of the corresponding feature, 
not its strength. The encoder weights and bias are rescaled accordingly:

\begin{align*}
   W^{dec}_{:,i} &\leftarrow \frac{W^{dec}_{:,i}}{\lVert W^{dec}_{:,i} \rVert_2},\\
   W^{enc}_{i,:} &\leftarrow \lVert W^{dec}_{:,i} \rVert_2 \cdot W^{enc}_{i,:},\\
   b^{enc}_i &\leftarrow \lVert W^{dec}_{:,i} \rVert_2 \cdot b^{enc}_i
\end{align*}

After these two post-processing steps, the SAE operates on the original input and attempts 
to reconstruct the original output. The decoder columns, now with unit 2-norm, represent 
the direction of each feature, while the hidden layer encodes the magnitude of the corresponding feature.

\section{Evaluation}\label{sec:evaluation}

We evaluate all our SAEs after post-training processing (Section~\ref{sec:post-training}), 
focusing primarily on two key aspects: how our SAEs advance the Pareto frontier of L0 sparsity and MSE efficiency
(Section~\ref{sec:pareto}), and the interpretability of the features extracted
with both automated and manual analysis (Section~\ref{sec:interpretability}). 

While our main focus is on these two areas, there are several other potential evaluation approaches. 
These include assessing the causal relevance of features to one another, as explored in circuit 
analysis~\citep{he2024othellocircuits, ge2024hierattr, dunefsky2024transcoders}, conducting 
fine-grained causal analysis with counterfactuals~\citep{Huang24ravel}, and performing automated 
prompt-based evaluations across a wide range of tasks~\citep{anthropic24aug}. We plan to explore 
these and emerging methods as the field of SAE evaluation continues to develop.

Additionally, we investigate whether the features learned by state-of-the-art SAE variants differ 
from those learned by vanilla SAEs (Section~\ref{sec:feature-geometry}). We also assess
the out-of-distribution generalization of Llama Scope SAEs across sequence length and to instruct-finetuned
models (Section~\ref{sec:ood}).

\subsection{Data and Metrics}

We evaluate the SAEs using three primary metrics: L0-norm of SAE latents, explained variance, and Delta LM loss. 
These metrics assess sparsity, reconstruction quality and the impact of the SAEs on the language model's 
performance, respectively.

Let $\hat{x}$ be the SAE's reconstruction of the input $x$, derived from Equation~\ref{eq:vanilla_decoder}
and~\ref{eq:encoder-our-topk}. The explained variance (EV) measures the proportion of variance in the input 
captured by the SAE and is defined as:

\begin{align}
   \text{EV} = 1 - \frac{\lVert x - \hat{x} \rVert_2^2}{\lVert x \rVert_2^2}.
\end{align}

In addition to reconstruction quality, we measure the impact of the SAE on the language model with Delta 
LM loss. This metric is defined as the difference between the original language model loss and the loss 
when the SAE is inserted at the corresponding position. Unlike explained variance and reconstruction MSE, 
which focus solely on the accuracy of the SAE’s reconstruction, Delta LM loss reflects the effect of the 
SAE on overall language model performance.

All evaluations are performed over 50 million tokens of held-out text data, sampled from the same 
distribution as the training data.

\subsection{Sparsity-Fidelity Pareto Efficiency}\label{sec:pareto}

\paragraph{TopK SAEs Outperform Vanilla SAEs.}

\begin{figure}[h]
   \centering

   \begin{subfigure}[b]{0.3\textwidth}
       \centering
       \includegraphics[width=\linewidth]{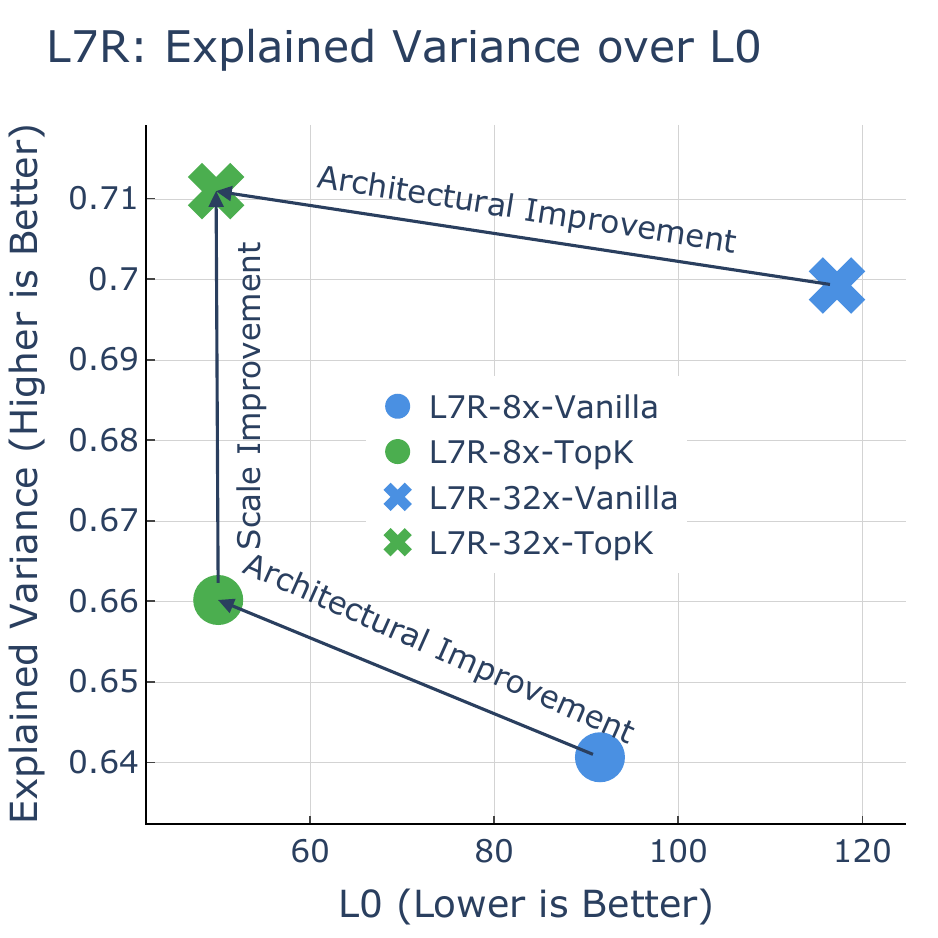}
   \end{subfigure}
   \hfill
   \begin{subfigure}[b]{0.3\textwidth}
       \centering
       \includegraphics[width=\linewidth]{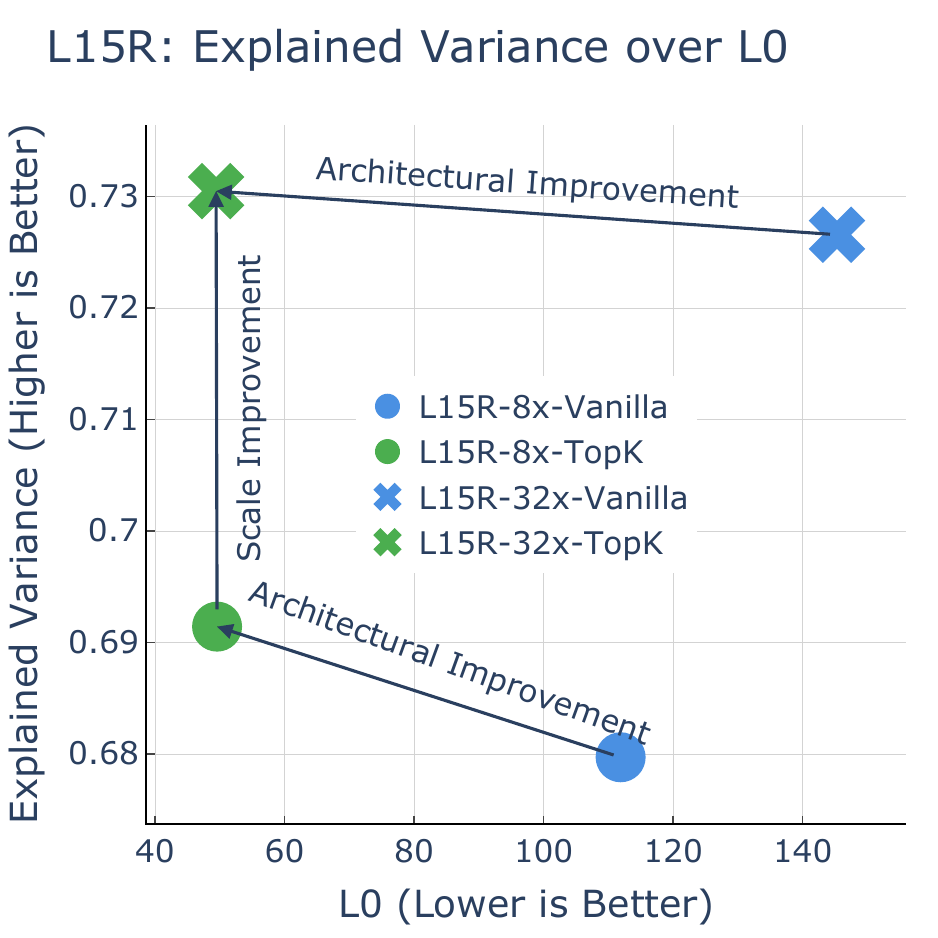}
   \end{subfigure}
   \hfill
   \begin{subfigure}[b]{0.3\textwidth}
       \centering
       \includegraphics[width=\linewidth]{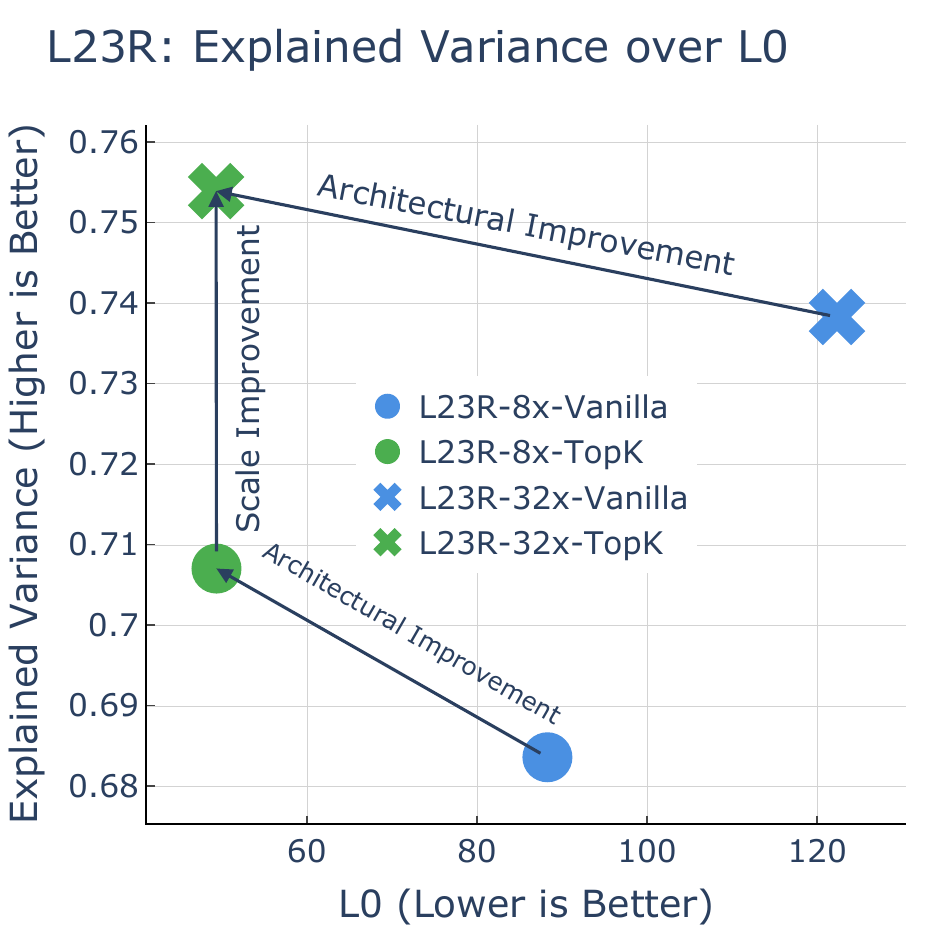}
   \end{subfigure}
   \vspace{0.2cm} 
   \begin{subfigure}[b]{0.3\textwidth}
       \centering
       \includegraphics[width=\linewidth]{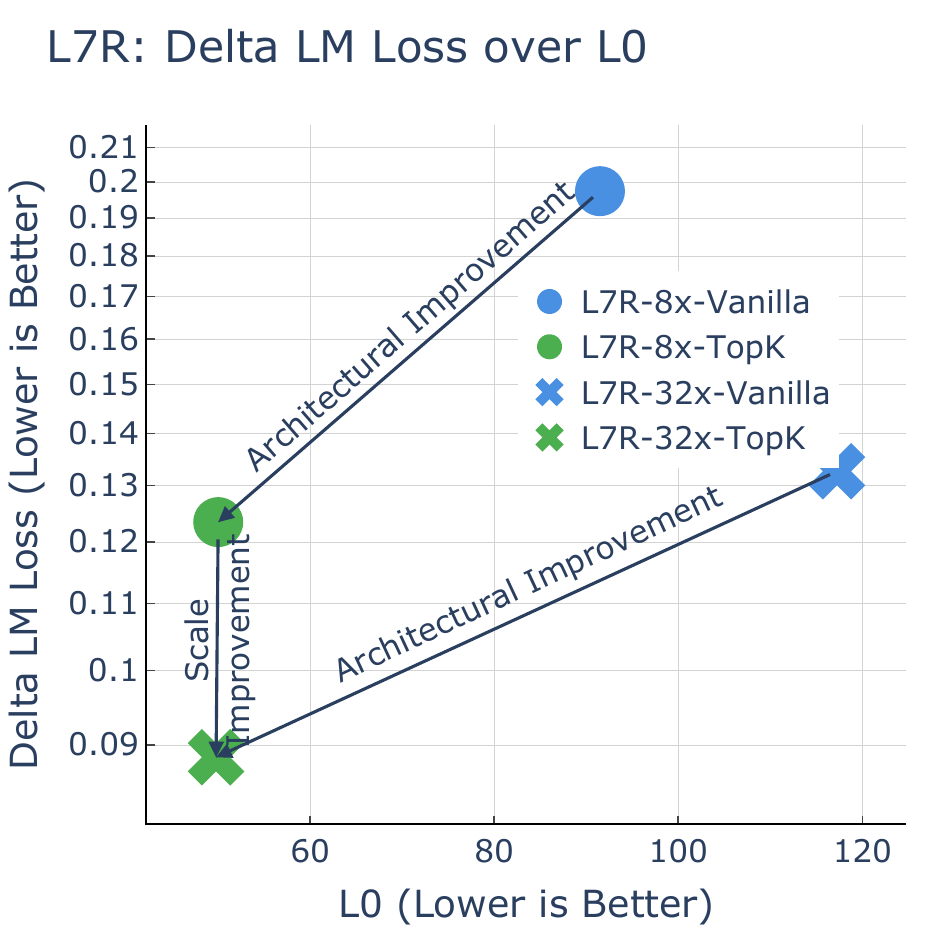}
   \end{subfigure}
   \hfill
   \begin{subfigure}[b]{0.3\textwidth}
       \centering
       \includegraphics[width=\linewidth]{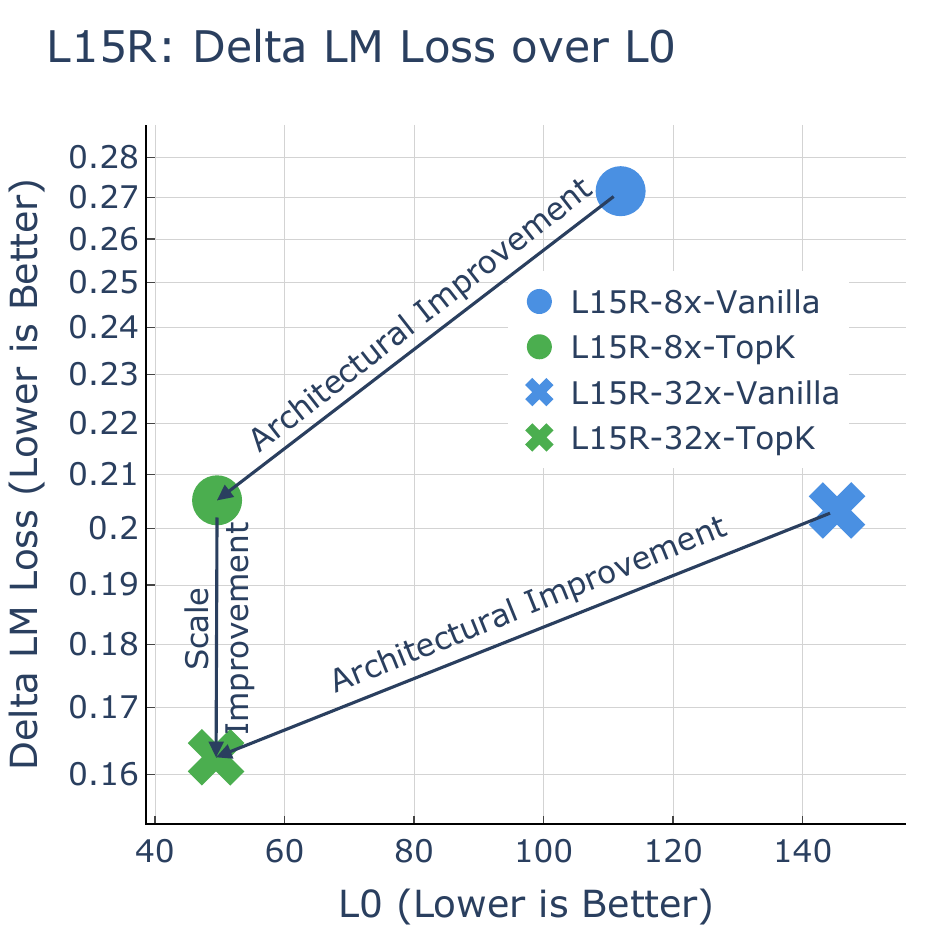}
   \end{subfigure}
   \hfill
   \begin{subfigure}[b]{0.3\textwidth}
       \centering
       \includegraphics[width=\linewidth]{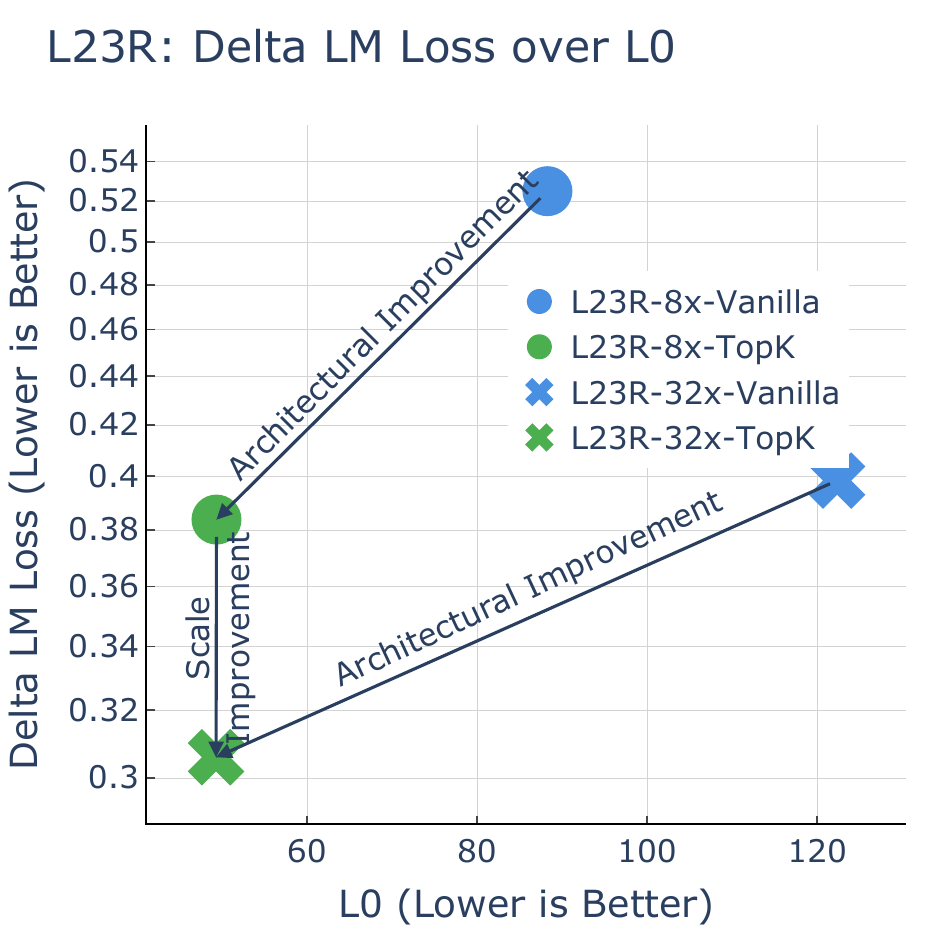}
   \end{subfigure}

   \caption{Explained Variance (upper) and Delta LM loss (lower) over L0 sparsity for SAEs trained on L7R, L15R and L23R.}
   \label{fig:ev_delta_ce_loss_over_L0_L7R_L15R_L23R}
\end{figure}

As shown in Figure~\ref{fig:ev_delta_ce_loss_over_L0_L7R_L15R_L23R}, TopK SAEs consistently outperform 
vanilla SAEs in both L0 sparsity and MSE efficiency when trained on each quartile layer's post-MLP 
residual stream. Holding all other hyperparameters constant, TopK SAEs reduce L0 sparsity from around 
150 to 50, while maintaining or improving both explained variance and Delta LM loss. This improvement 
is likely due to TopK SAEs (1) mitigating the feature shrinkage issue, and (2) removing weakly firing 
features.

\paragraph{Wider SAEs Achieve Better Pareto Efficiency.}

Figure~\ref{fig:ev_delta_ce_loss_over_L0_L7R_L15R_L23R} also shows that wider SAEs
outperform narrower ones in reconstruction while maintaining the same L0 sparsity. 
This suggests that, based on these high-level metrics, there is a positive correlation between the number 
of features and the overall quality of the SAEs. However, it is possible that wider SAEs simply learn 
more frequent compositions of existing features rather than discovering entirely new 
ones~\citep{evan24composed}. We provide a counterexample to this in Section~\ref{sec:feature-geometry}.

\paragraph{Overall Assessment.}

Evaluation results for all 256 SAEs are illustrated in Figure~\ref{fig:overall eval} in Appendix~\ref{sec:eval_all}. Across all positions, layers and widths, TopK SAEs
achieve consistently match or slightly exceed the reconstruction quality of vanilla SAEs, while achieving significantly better L0 sparsity.
Wider SAEs show lower Delta LM loss and higher explained variance, all while maintaining the same L0 sparsity.

Residual stream SAEs (LXR-8x and LXR-32x) generally outperform those trained on other positions in terms of the 
aformentioned 3 metrics.
We conjecture this is due to Cross Layer Superposition~\citep{anthropic24july} where the output of each model component 
only form a small fraction to certain features, potentially introducing noise into the SAE training process.

\subsection{Interpretability of Features}\label{sec:interpretability}

\begin{figure}[h]
   \centering
   \includegraphics[width=\linewidth]{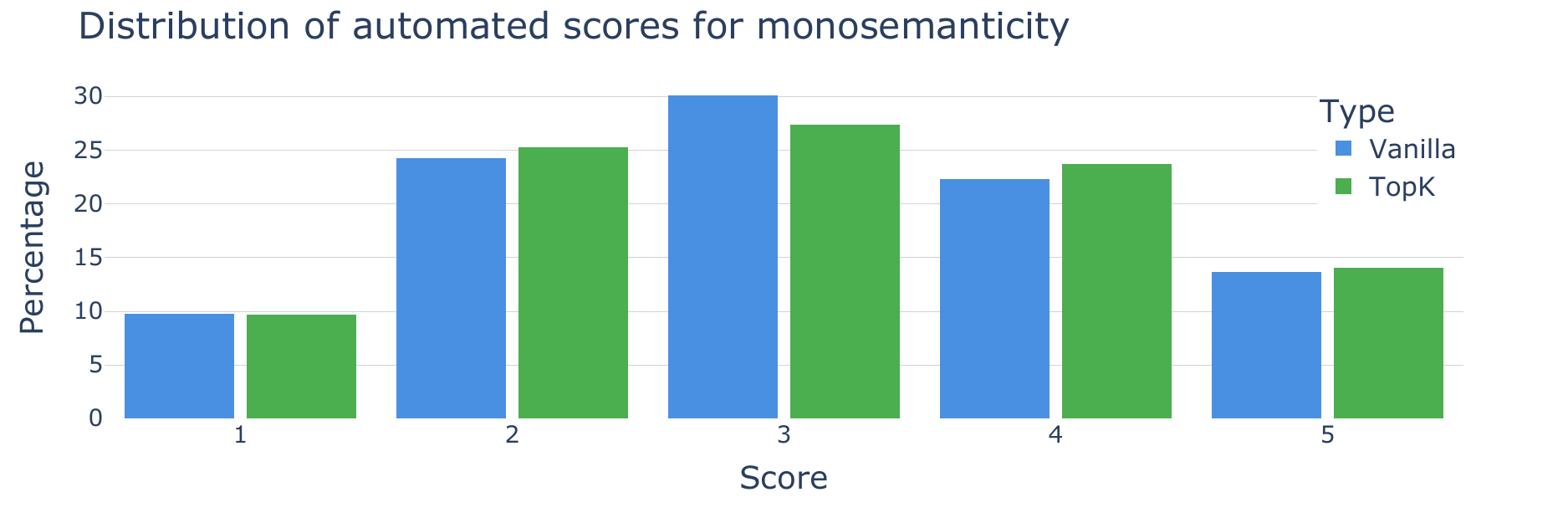}
   \caption{Automatically labeled monosemanticity scores of L15R-8x SAE features.}
   \label{fig:mono}
\end{figure}

We assess the interpretability of the features extracted by SAEs using automated analysis, following the 
approach of~\citet{Bills2023autointerp, anthropic24june}. For each feature, the 20 most activating contexts 
are sent to GPT-4o to generate a description of the feature. GPT-4o also scores each feature's monosemanticity, 
from 1 (Not comprehensible) to 5 (Clear and consistent pattern), using the rubric adapted from~\citet{anthropic24june}:

\begin{itemize}
   \item \textbf{5}: Clear pattern with no deviating examples;

   \item \textbf{4}: Clear pattern with one or two deviating examples;

   \item \textbf{3}: Clear overall pattern but quite a few examples not fitting that pattern;

   \item \textbf{2}: Broad consistent theme but lacking structure;

   \item \textbf{1}: No discernible pattern.
\end{itemize}

We randomly selected 128 features from the L7R-8x, L15R-8x, and L23R-8x SAEs, for both TopK and Vanilla SAEs, 
resulting in a total of 768 features. The monosemanticity results, shown in Figure~\ref{fig:mono}, indicate 
no significant difference between TopK and Vanilla SAEs.

Additionally, we performed manual analysis on a subset of these features. The manual evaluation is 
consistent with the automated analysis, finding that about 10\% of the features are not interpretable 
(monosemanticity score of 1). We did not further distinguish between scores of 2-5, as it is difficult 
to draw clear distinctions between these levels of interpretability.

It's important to note that our analysis focuses on top activations, which may not assess each feature's behavior
on low-activating samples. It has been observed that lower activating samples tend to be less relevant to 
the interpretation~\citep{bricken2023monosemanticity} for vanilla SAE features. Both TopK and JumpReLU 
SAEs are designed to remove weakly firing instances and prioritize features with higher activation 
levels, based on a threshold or rank-based thresholding. This may not reflected in the monosemanticity
scores, which is a limitation of our current evaluation.

\subsection{Activation Frequency}

The firing frequency of latent features in SAEs is a crucial metric for validating the correctness of the 
training process and diagnosing potential issues early. If too many features fire frequently, they are 
likely to be uninterpretable. Conversely, if features rarely fire across a large number of tokens, it 
could indicate that the sparsity constraints are too strict.

We empirically find that a satisfying proportion of inactive features (those that do not fire once over 
100,000 tokens) is around 10\%. This includes both ultra-rare features and potential training failures. 
For larger SAEs, this threshold may be higher, but we use it to guide the training of Llama Scope SAEs. 
Additionally, we aim to keep ultra-active features (those firing with a frequency greater than 0.1) below 
2\% of all features.

\begin{figure}[h]
   \centering
   \begin{subfigure}[b]{0.31\textwidth}
      \centering
      \includegraphics[width=\linewidth]{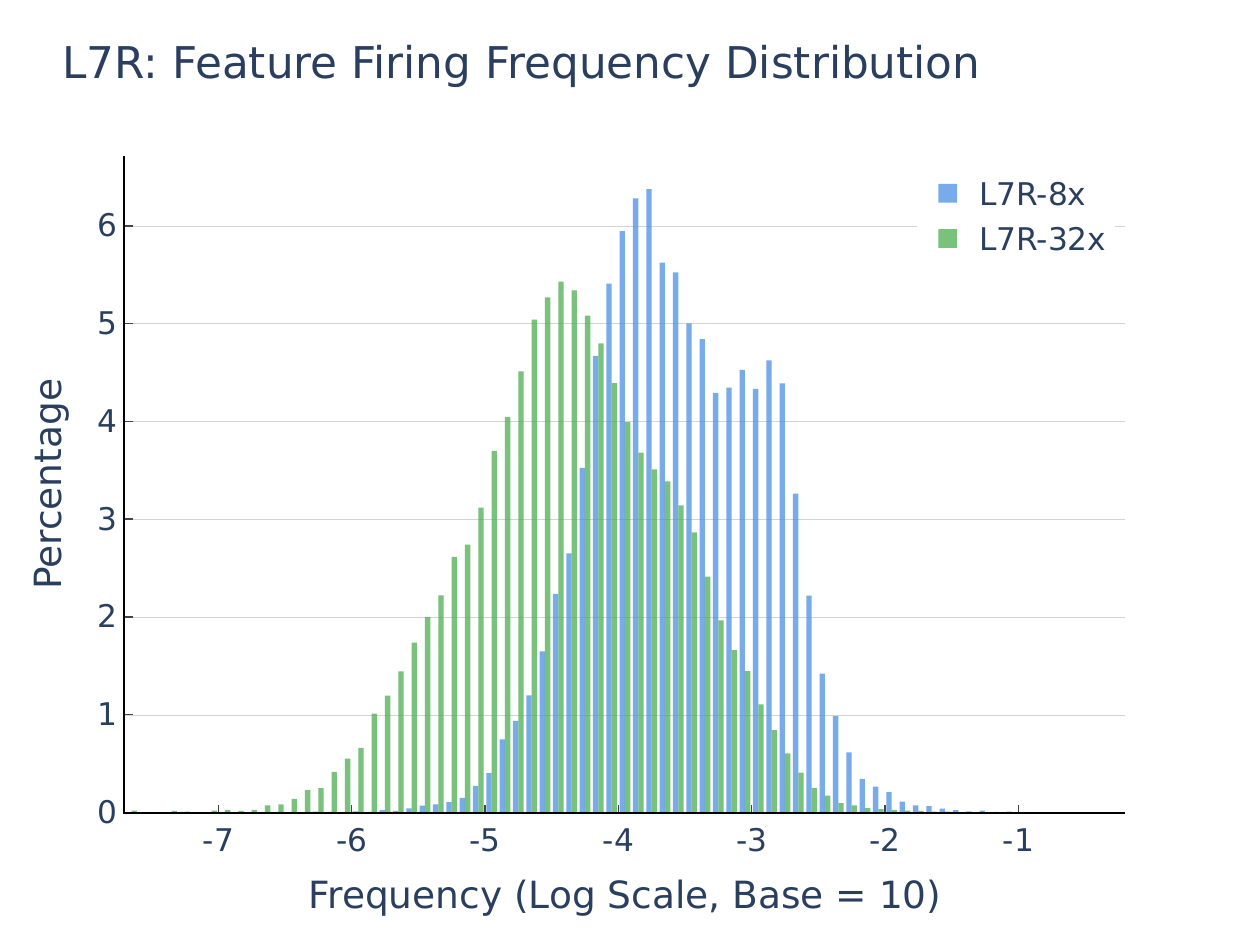}
  \end{subfigure}
  \begin{subfigure}[b]{0.31\textwidth}
      \centering
      \includegraphics[width=\linewidth]{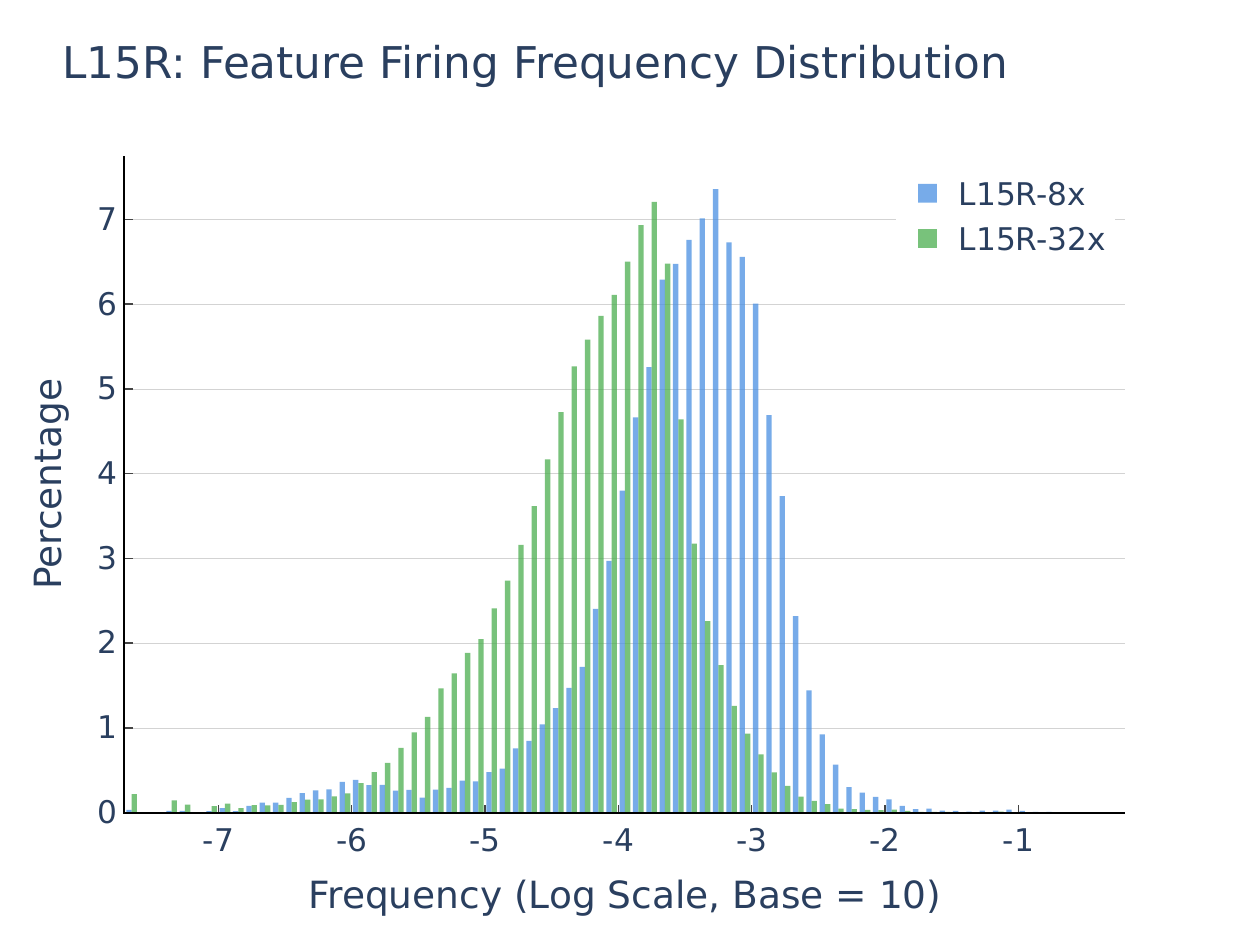}
  \end{subfigure}
  \begin{subfigure}[b]{0.31\textwidth}
      \centering
      \includegraphics[width=\linewidth]{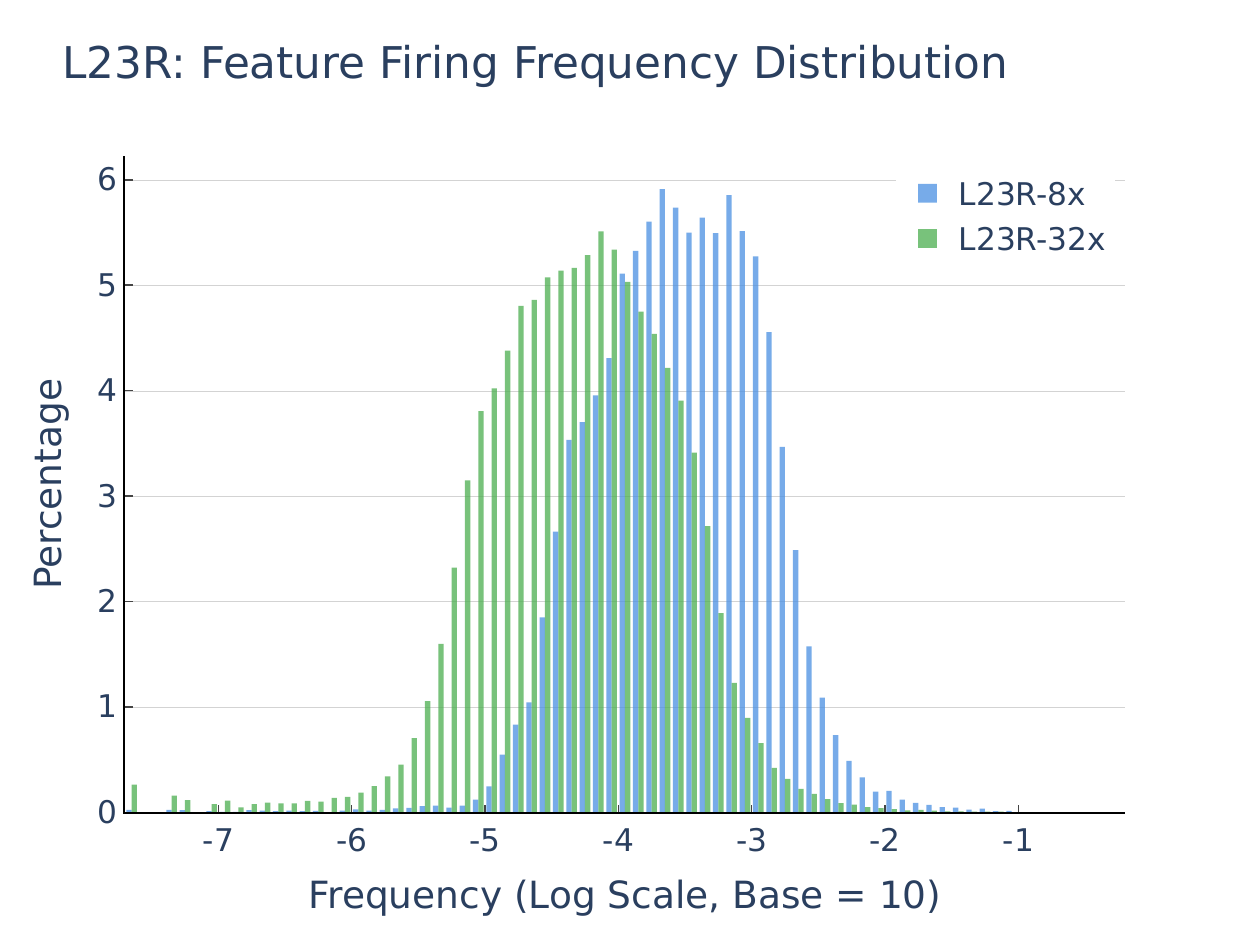}
  \end{subfigure}
  \caption{Firing frequency of L7R-8x, L15R-8x and L23R-8x TopK SAEs.}\label{fig:firing_freq}
\end{figure}

We show the firing frequency of L7R-8x, L15R-8x and L23R-8x TopK SAEs in Figure~\ref{fig:firing_freq}.
The firing frequency distribution of wider SAEs tend to \emph{left-shift} towards lower frequency 
compared to narrower ones.
This is expected, as wider SAEs have more features whereas top-k sparsity constraint is fixed at 50
for both of them.

\subsection{Out-of-Distribution Generlization}\label{sec:ood}

Since SAE training is resource-intensive, it's important that the models generalize beyond the 
training distribution. We evaluate the out-of-distribution generalization of our SAEs in two ways: 
on longer contexts and on instruction-finetuned models.

\subsubsection{Across Sequence Length}

\begin{figure}[h]
   \centering
   \begin{subfigure}[b]{0.48\textwidth}
      \centering
      \includegraphics[width=\linewidth]{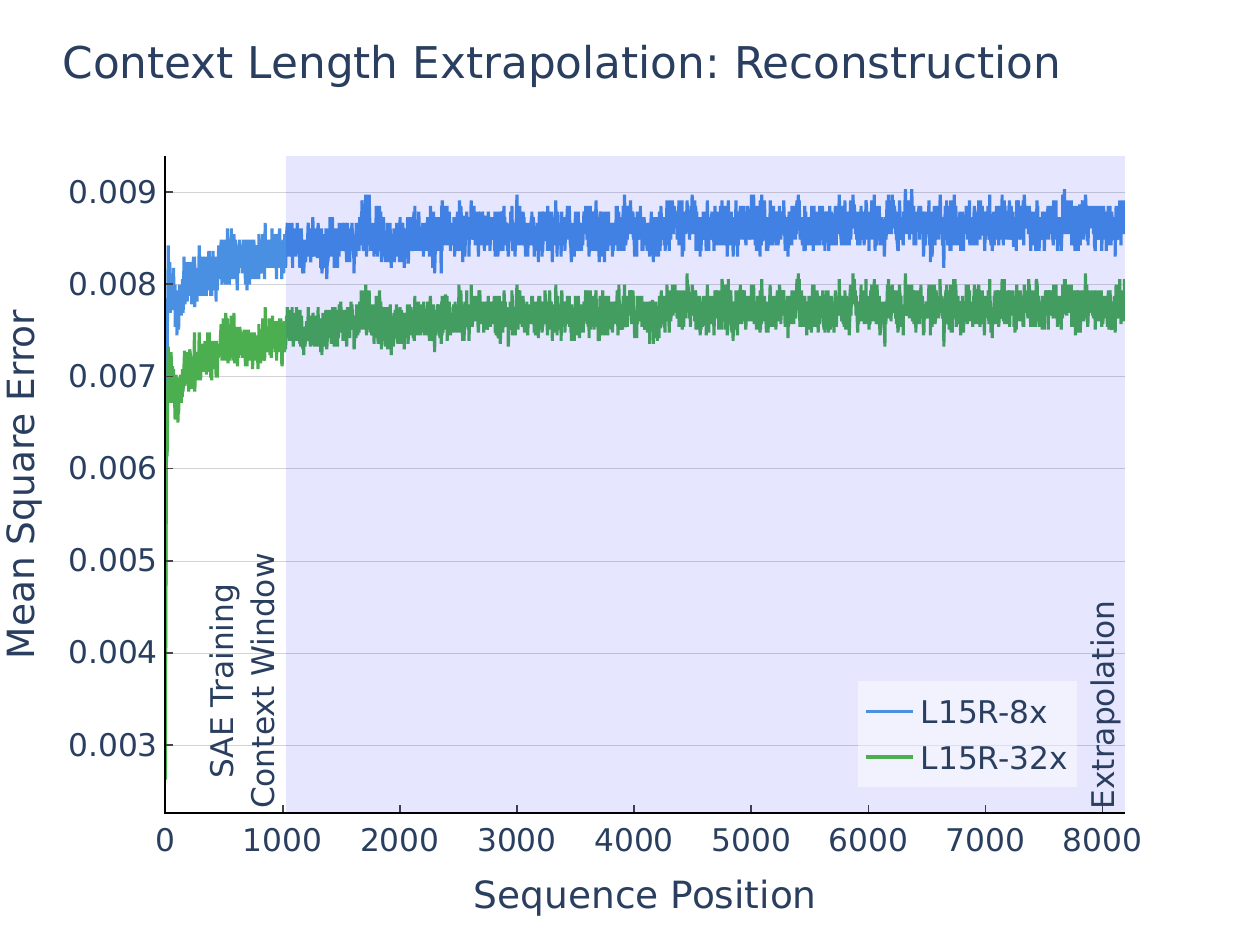}
  \end{subfigure}
  \begin{subfigure}[b]{0.48\textwidth}
      \centering
      \includegraphics[width=\linewidth]{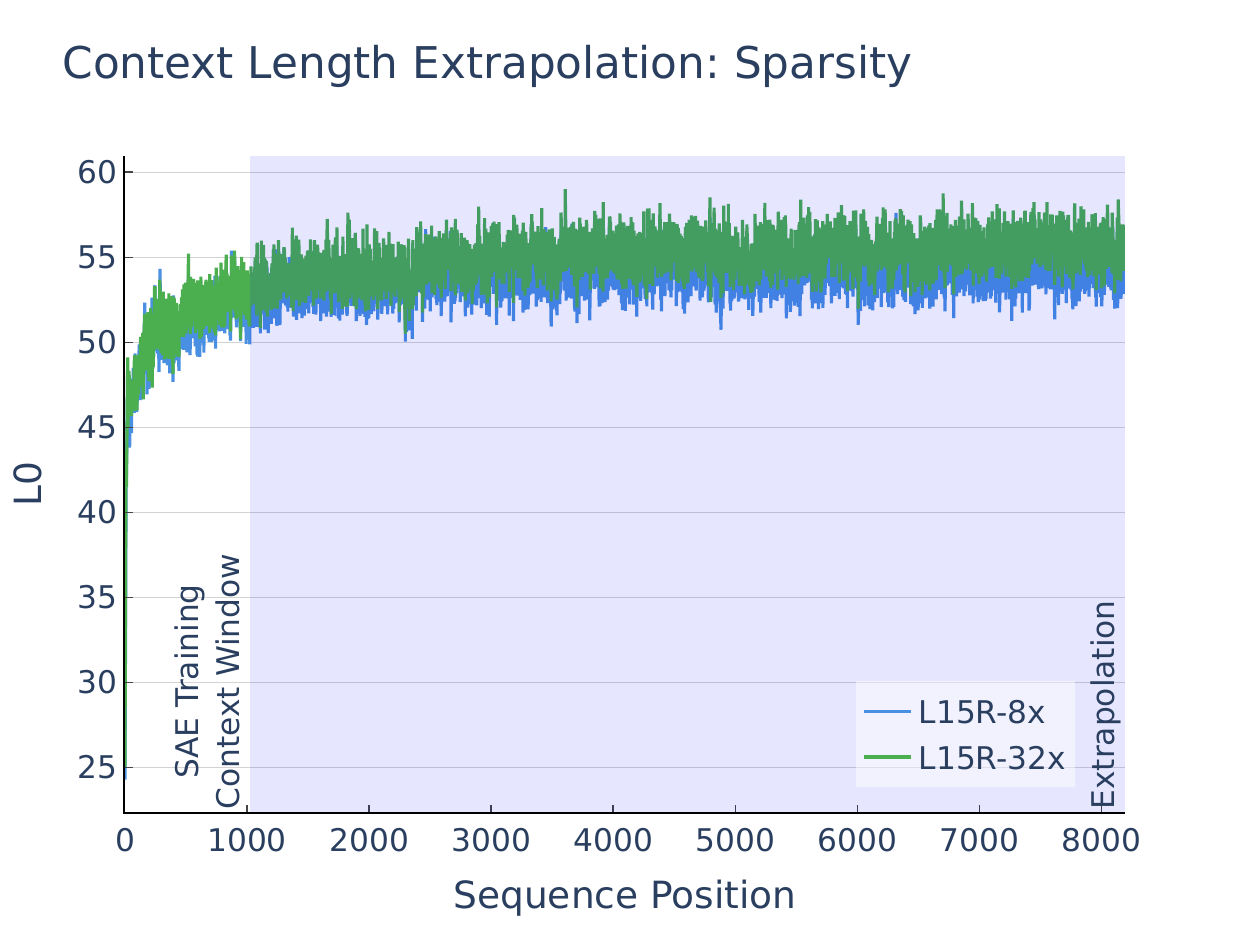}
  \end{subfigure}
  \caption{SAE performance on long context data, measured by MSE and L0 sparsity.}\label{fig:ood_length}
\end{figure}

We selected all documents with more than 8192 tokens from L-Eval~\citep{an23leval} and truncated them to
8192 tokens, resulting in a dataset of 166 documents and 1.3 million tokens. The L0 sparsity and MSE of 
Llama Scope SAEs on this dataset are shown in Figure~\ref{fig:ood_length}.

Since our SAEs are trained on 1024-token sequences (Section~\ref{sec:collect_activations}), a slight 
decrease in performance with longer sequences is expected. However, the degradation
converges at around 8192 tokens. The average reconstruction loss in the last 1024 tokens 
is 0.0086, which is a 12\% increase compared to the training data. The L0 sparsity increases 
from 50 to 55. 

Although we have not fine-tuned our SAEs on longer sequences, we believe they can be further optimized for 
long-context data with minimal fine-tuning, which we plan to explore in future work.

\subsubsection{To Instruction-Finetuned Models}

\begin{figure}[h]
   \centering
   \includegraphics[width=\textwidth]{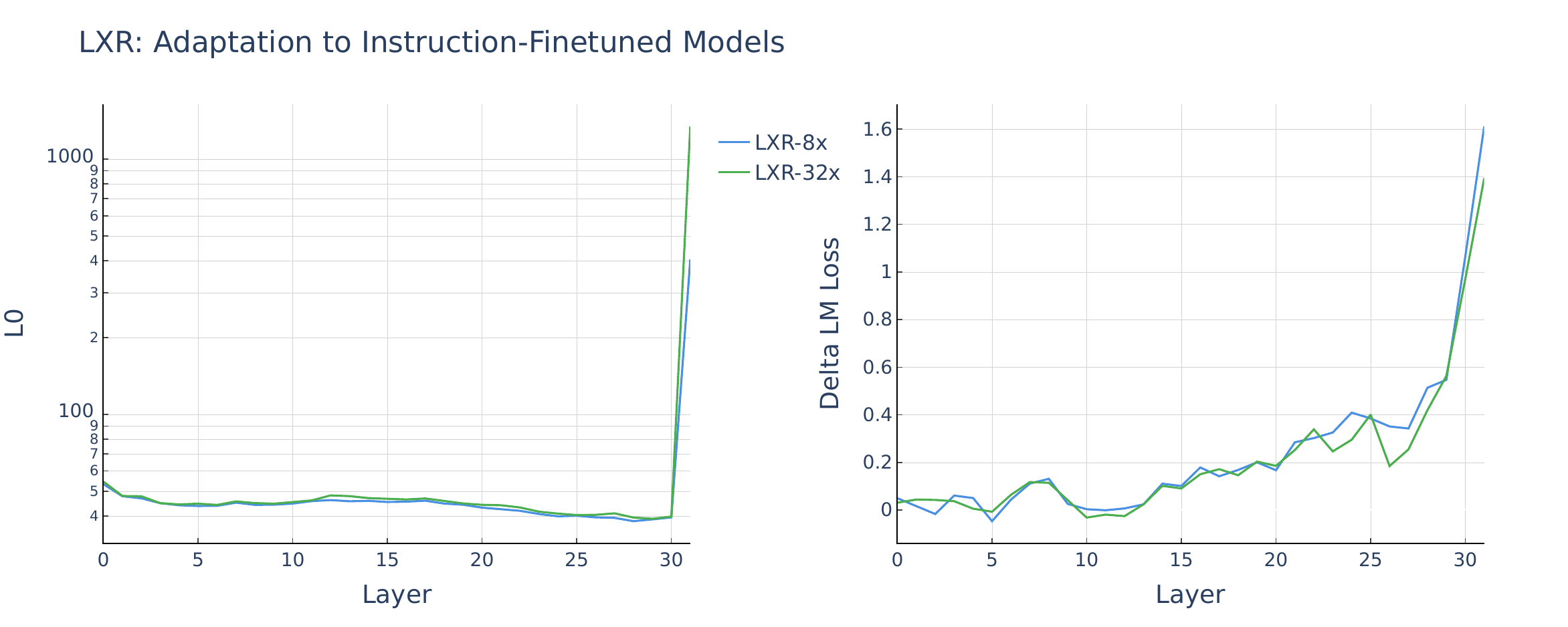}
   \caption{L0 sparsity and Delta LM loss of L15R-8x TopK SAEs on instruction-finetuned models.}\label{fig:ood_it}
\end{figure}

When discussing the generalization of SAEs to instruction-finetuned models, there are three key dimensions to 
consider: the language model, SAE training data, and downstream tasks.

In~\citet{templeton2024scaling}, SAEs trained on pretraining data with an instruction model generalized well 
to downstream instruction-guided tasks\footnote{Interestingly, they also showed generalization to image data.}. 
This can be described as \emph{base SAE training data generalizing to chat tasks} while keeping the model 
fixed to instruction-finetuned models.

Our setup is similar to that of~\citet{kissane24chattransfer}, where \emph{base SAE training data and a base 
language model are generalized to chat models and chat data}. Specifically, we use a subset of the Wildchat 
dataset~\citep{zhao24wildchat} with 4K chat histories. These are fed into Llama-3.1-8B-Instruct to generate 
activations, and we evaluate 32 LXR-8x-TopK and 32 LXR-32x-TopK SAEs on this dataset~\ref{fig:ood_it}.

With the exception of L31R-8x and L31R-32x (trained on residual stream activations just before the final 
layernorm), no significant degradation in Delta LM loss or increase in L0 sparsity is observed across the 
other SAEs\footnote{The exceptions in the near-output space may offer interesting insights, though we do 
not explore them further here.}. This indicates that our SAEs generalize well to instruction-finetuned models.

For example, replacing L15R activations with those reconstructed by L15R-32x-TopK SAEs in the base model 
increases the language modeling loss by 0.162, whereas the increase is only 0.090 for instruction-finetuned 
models, all while maintaining the same L0 sparsity. The smaller Delta LM loss in Llama-3.1-8B-Instruct does 
not necessarily mean that the SAEs are better suited for instruction-finetuned models, but rather that the 
downstream tasks are more robust to the SAE's perturbations.

\subsection{Feature Geometry}\label{sec:feature-geometry}

To better understand the behavior of SAEs across different architectures and sizes, we analyze their feature 
geometry, following the approach of~\citet{bricken2023monosemanticity, templeton2024scaling}. Specifically, 
we examine the cosine similarity between decoder columns $W^{dec}_{:,i}$ to identify adjacent features in 
the activation space (the \emph{decoder} view). For this analysis, we selected a feature in L15R-8x-TopK 
that activates in contexts related to 
significant historical events, particularly those harmful to humanity. We then found the six most similar 
features in both L15R-8x-Vanilla and L15R-8x-TopK, and the 24 most similar features in L15R-32x-TopK.

It is important to note that the cosine similarity between two features is not a direct measure of their
semantic similarity. However, we do find that features with high cosine similarity tend to have similar
interpretations, as shown in the following section.

\begin{figure}[h]
   \centering
   \includegraphics[width=\linewidth]{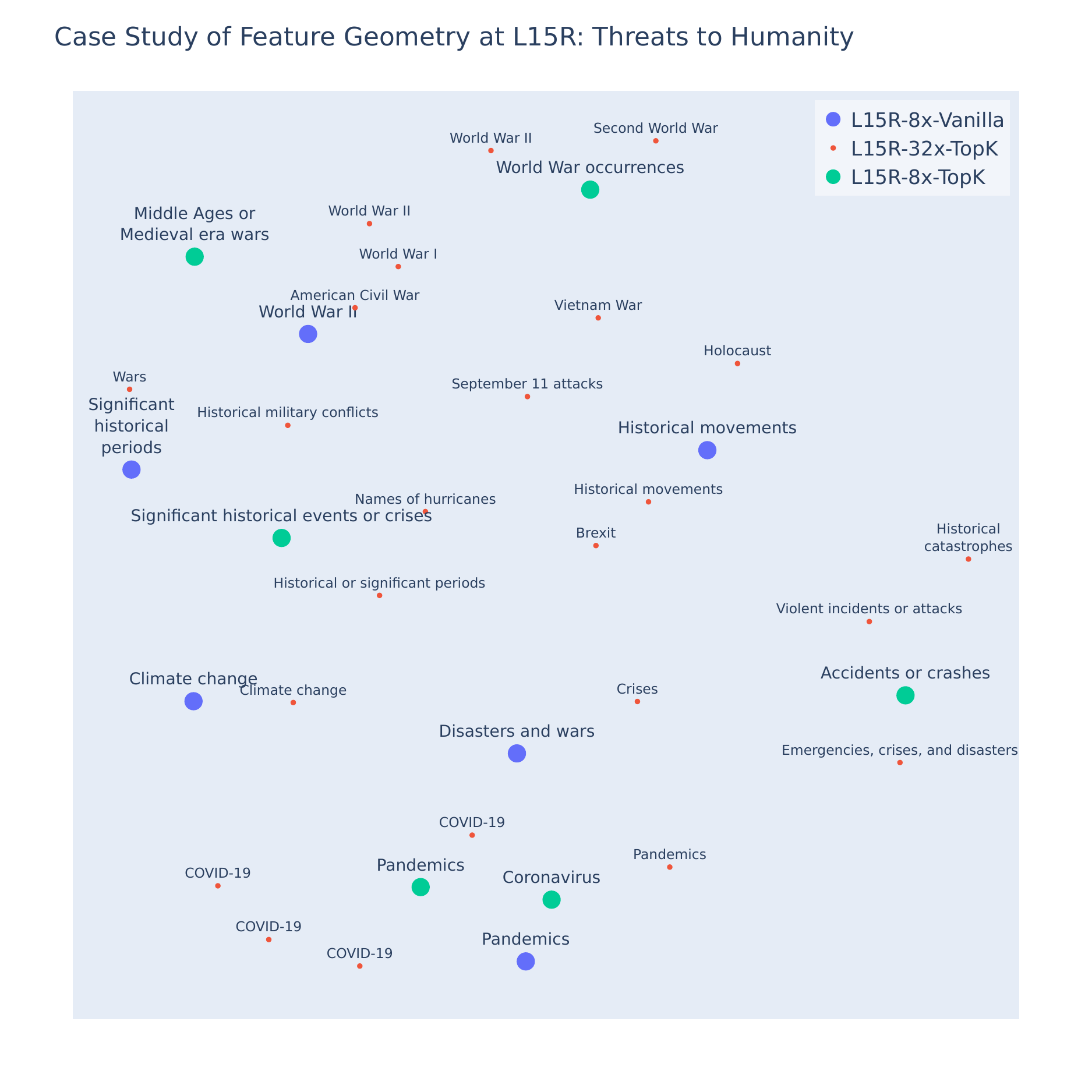}
   \caption{Feature geometry of L15R-8x-Vanilla, L15R-8x-TopK and L15R-32x-TopK SAEs.}\label{fig:feature_geometry}
\end{figure}

\paragraph{The Threats-to-Humanity Cluster.} Figure~\ref{fig:feature_geometry} shows the 2D UMAP projection 
of the decoder columns for these 36 features. Using GPT-4o to describe their top activating samples (the 
\emph{encoder} view), we found a coherent theme across all features: they are related to events like wars, 
pandemics, natural disasters, and other harmful occurrences.

UMAP, which preserves the local structure of data, also reveals more detailed local clusters. For instance, 
the top-left region of Figure~\ref{fig:feature_geometry} contains features associated with large-scale wars, 
while other regions correspond to themes such as \emph{climate change}, \emph{pandemics}, and \emph{financial crises}.

\paragraph{How Close Enough are Features to Be Neighbors?} 
One challenge in identifying \emph{feature neighbors} 
is the lack of a principled threshold for determining whether two features are close enough to be considered neighbors. 
To address this, we provide a baseline based on the Johnson-Lindenstrauss lemma's inner product version. If we randomly 
project $F$ \emph{true} features to the $D$-dimensional hidden space, the probability
of finding a pair (out of $F\choose 2$ pairs) that have cosine similarity larger than
$\epsilon=\sqrt{\frac{12\ln F}{D}}$ is approximately $2/F$.\footnote{Proof can be found at
\url{https://home.ttic.edu/~gregory/courses/LargeScaleLearning/lectures/jl.pdf}. The small difference between inner product
and cosine similarity is neglected for estimation purposes.}

For 8x SAEs, $\epsilon = 0.174$, and for 32x SAEs, $\epsilon = 0.186$. We empirically validate this by randomly 
mapping $F = 8 \times 4096$ (or $32 \times 4096$) features into a $D = 4096$-dimensional space, where we find 
the maximum pairwise cosine similarity to be 0.09 (0.10). In comparison, the farthest cosine similarity distance 
between the 36 features in Figure~\ref{fig:feature_geometry} is 0.28, suggesting that these features exhibit 
non-trivial similarity.

\paragraph{Wider SAEs Do Learn New Features.} In this case, features from L15R-32x-TopK are not merely linear 
combinations of features from L15R-8x-TopK. For example, there is a distinct \emph{Brexit} feature in 
L15R-32x-TopK that activates exclusively on this topic, whereas the most similar feature in the smaller 
SAEs is a more general \emph{Historical Movements} feature. This suggests that wider SAEs are capable of 
learning entirely new features, not just frequent compositions of existing ones.

\paragraph{TopK SAEs Learn Similar Features to Vanilla SAEs.} We also observe that the features learned 
by TopK SAEs closely resemble those in Vanilla SAEs, as measured by cosine similarity of the decoder columns. 
This is visually supported by the fact that the blue dots (L15R-8x-Vanilla) and green dots (L15R-8x-TopK) are 
intermixed in Figure~\ref{fig:feature_geometry} and share a common theme.

\begin{figure}[ht]

   \centering
   \includegraphics[width=\linewidth]{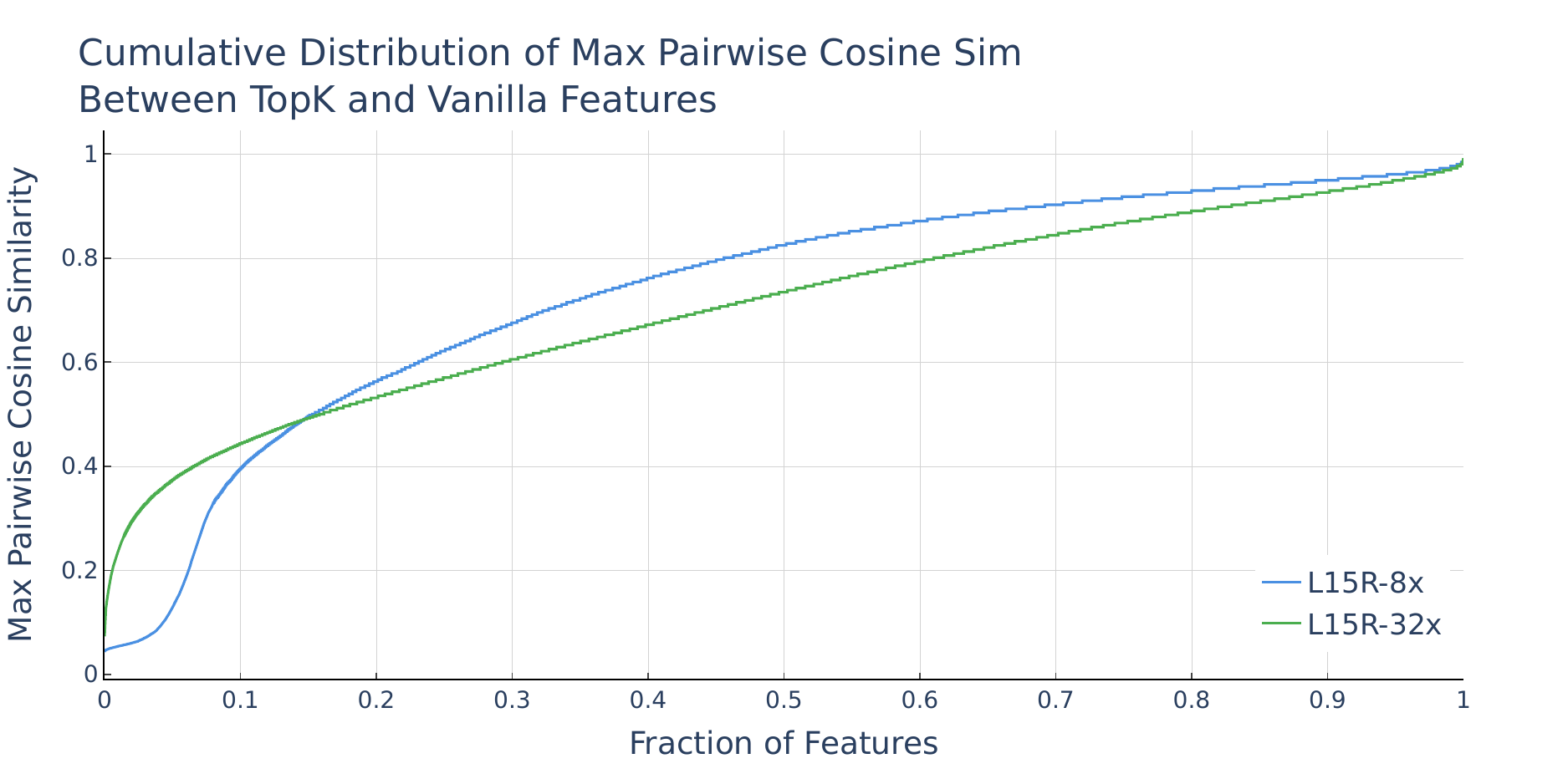}
   \caption{Feature geometry of L15R-8x-Vanilla, L15R-8x-TopK and L15R-32x-TopK SAEs.}\label{fig:global_similarity}
\end{figure}

For a broader analysis, we computed the cosine similarity between all features in L15R-8x-TopK and their most 
similar counterparts in L15R-8x-Vanilla. Figure~\ref{fig:global_similarity} shows the cumulative distribution 
of these similarity scores. The similarity between TopK and Vanilla SAEs is significantly higher than a random 
baseline (where the max cosine similarity between two random SAEs of the same size is computed). This indicates 
that TopK and Vanilla SAEs share a universal feature geometry.

\section{Related And Future Work}\label{sec:related-work}

\paragraph{Scaling up Sparse Autoencoders.} Building on recent success in scaling SAEs to large language 
models~\citep{templeton2024scaling, lieberum2024gemmascope, gao2024oaisae}, further scaling is likely to continue 
with larger models. The development of customized GPU acceleration for SAEs~\citep{gao2024oaisae} holds great potential 
for extracting more complex concepts from larger models~\citep{kaplan20scalinglaw, hoffman22chinchillascalinglaw}. 
Additionally, structured sparsity in Mixture-of-Expert (MoE) SAEs, as mentioned in~\citet{gao2024oaisae} 
and~\citet{sharkey24ideas}, has shown promise for efficient SAE training~\citep{mudide2024switchsaes}. This approach 
could significantly reduce computational costs while maintaining performance, making it a promising avenue for 
future exploration.

\paragraph{Extending Neuron-Level Analysis to SAE Features.} SAE features share similarities with MLP neurons, which 
have been widely studied for their role in knowledge representation~\citep{dai22knowledgeneurons}, 
universality~\citep{gurnee2023sparseprobing, gurnee24univ}, and multilingual capabilities~\citep{tang24multilingualneurons}. 
Insights from these studies could be applied to SAE features to deepen our understanding of language models. Recent 
work~\citep{wang2024univ} suggests that universality across model architectures can be reflected more accurately by 
SAE features than by MLP neurons.

\paragraph{Revealing a More Interpretable Latent Space.} Understanding the activation space of language models is a 
central challenge in interpretability research. We believe techniques used in language model activation spaces can 
be applied to SAEs' latent spaces. For instance, linear probes and decision tree classifiers have been successfully 
used to predict harmful content from SAE features~\citep{bricken2024saeprobe}. Additionally, similarity metrics like 
Canonical Correlation Analysis (CCA) have been shown to provide valuable insights into SAE latent spaces~\citep{lan2024sparse}.

\section{Conclusion}
In this work, we introduced Llama Scope SAEs, a series of TopK Sparse Autoencoders trained on the Llama-3.1-8B-Base 
model. The insights and lessons learned from this work provide a valuable foundation for future improvements in SAE 
training. As mechanistic interpretability remains an open field with many unexplored ideas, we hope that Llama Scope, 
alongside other open-source SAEs, will help researchers save time and effort in their investigations.

Both language models and SAEs are evolving rapidly, and we see this as a long-term research direction. We are 
excited to contribute to the growing body of open-source SAEs and to continue pushing the epistemic frontier of 
SAEs in the future.

\newpage

\bibliography{iclr2025_conference}
\bibliographystyle{iclr2025_conference}

\newpage

\appendix

\section{Infrastructure}\label{sec:infrastructure}

\subsection{Activation Buffer}\label{sec:activation-buffer}

One of the major challenges in training SAEs is the substantial storage and throughput 
required for latent activations. While text data requires only 2 bytes per token, latent 
activations occupy 8K bytes per token—resulting in a 4,096x increase in both storage 
needs and disk throughput. This, combined with the relatively fast training steps of 
shallow SAEs, means that data loading quickly becomes the main bottleneck in the training process.

Due to these infrastructure constraints, we do not save activations in advance but instead 
generate them on-the-fly. This contrasts with the approach taken 
by~\citet{lieberum2024gemmascope, templeton2024scaling}, where activations are pre-saved 
and a high-speed dataloading pipeline is built to keep up with training.

To manage this, we adopt a producer-consumer model. Language Models (LMs) generate 
activations and store them in an activation buffer, while the SAEs consume the activations 
in random order. The process is serialized: once the buffer is full, SAE training begins, and 
when half the buffer is consumed, the LMs refill it. Each time the buffer is refilled, we 
shuffle it to introduce randomness into the training data without needing to save and 
shuffle all activations at once.

\subsection{Mixed Parallelism}

\begin{figure}[h]
   \begin{minipage}{0.4\textwidth}
       \centering
       \includegraphics[width=0.99\linewidth]{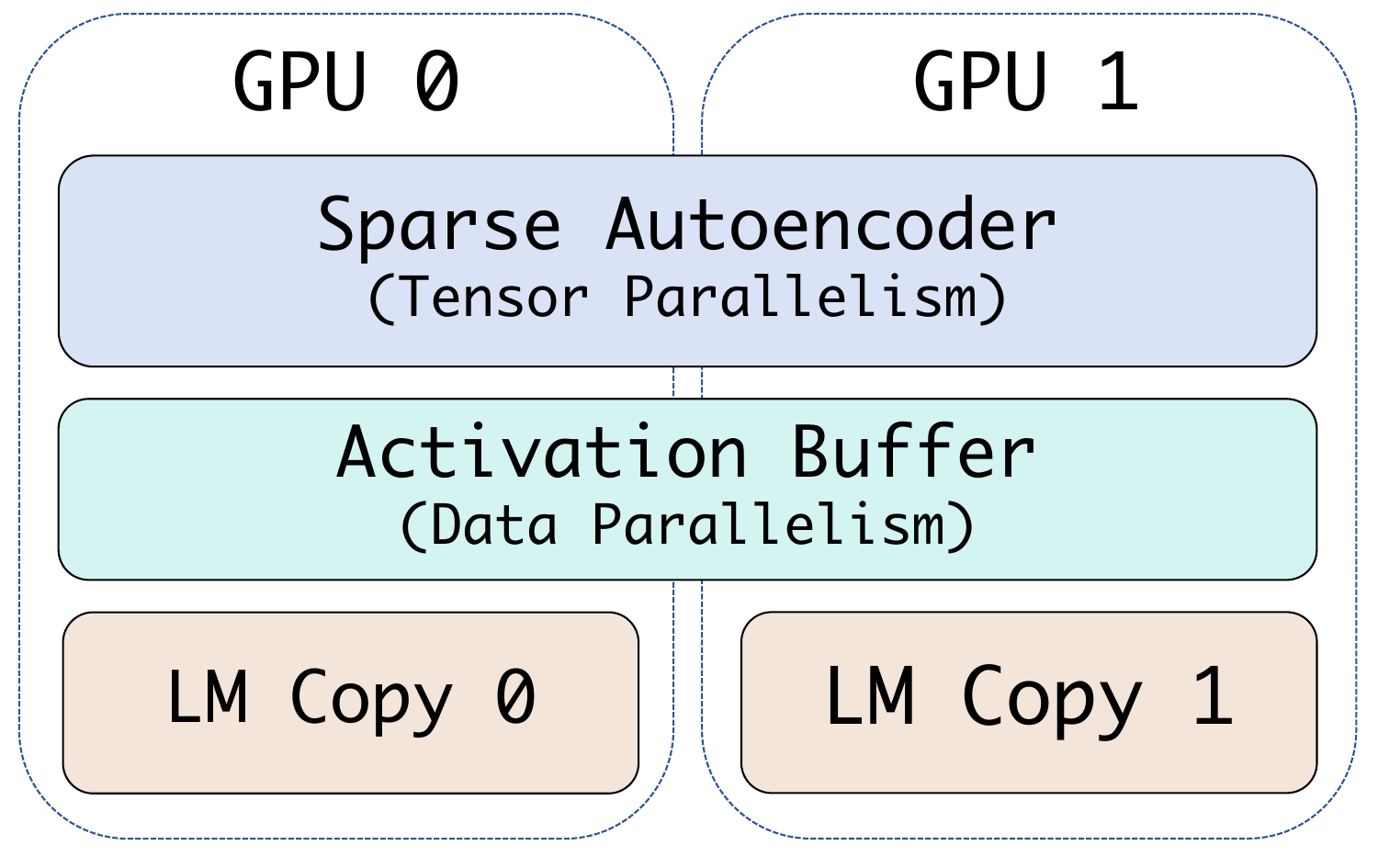}
       \caption{Mixed parallelism strategy for training SAEs.}\label{fig:mixed-parallelism}
   \end{minipage}
   \hfill 
   \begin{minipage}{0.56\textwidth}
      To efficiently train SAEs with a large number of features, we implement a mixed 
      parallelism strategy, which balances the memory demands of SAE training with the 
      slower process of activation generation. This approach accelerates buffer refilling 
      and reduces the memory bottleneck during training.
      
      As shown in Figure~\ref{fig:mixed-parallelism}, a copy of the Language Model (LM) 
      is loaded on each GPU, while the SAE is distributed across GPUs using tensor 
      parallelism. Each GPU maintains its own independent activation buffer. During 
      training, a minibatch is sampled from the activation buffer on each GPU, and these 
      minibatches are all-gathered to form the input for the SAE.
   \end{minipage}
\end{figure}

This setup combines data parallelism for activation generation with tensor parallelism 
for SAE training. We empirically find that SAEs and their gradients consume significant 
memory, but individual training steps are relatively fast. In contrast, activation 
generation is slower but requires less memory. By combining these parallelism techniques, 
mixed parallelism accelerates activation generation while efficiently managing memory 
during SAE training.

\subsection{Comparison to Pre-Saving Approach}

Previous work on training large SAEs using distributed disk reading and extensive storage 
resources~\citep{lieberum2024gemmascope, templeton2024scaling} has typically been conducted 
in industrial labs with powerful infrastructure. In contrast, our online activation 
generation approach eliminates the need for vast storage resources, making it more suitable 
for academic research. With this method, SAEs can be trained on an 8B-size model using just 
a single NVIDIA A100 GPU, without the need to pre-save activations. In comparison, the 
pre-saving approach requires approximately 2TB of storage and a disk throughput of at least 
500MB/s to ensure that data loading keeps pace with training.

However, a key limitation of our online generation method is the potential for redundant 
computation when training multiple SAEs on the same position. If we need to train SAEs with 
different widths at the same position, the same activations must be generated multiple times, 
leading to 1-2 times the redundant activation generation compared to the pre-saving approach.

\section{Evaluation Result for All SAEs}\label{sec:eval_all}

\begin{figure}[bp]
   \centering

   \begin{subfigure}{\textwidth}
       \centering
       \includegraphics[width=\linewidth]{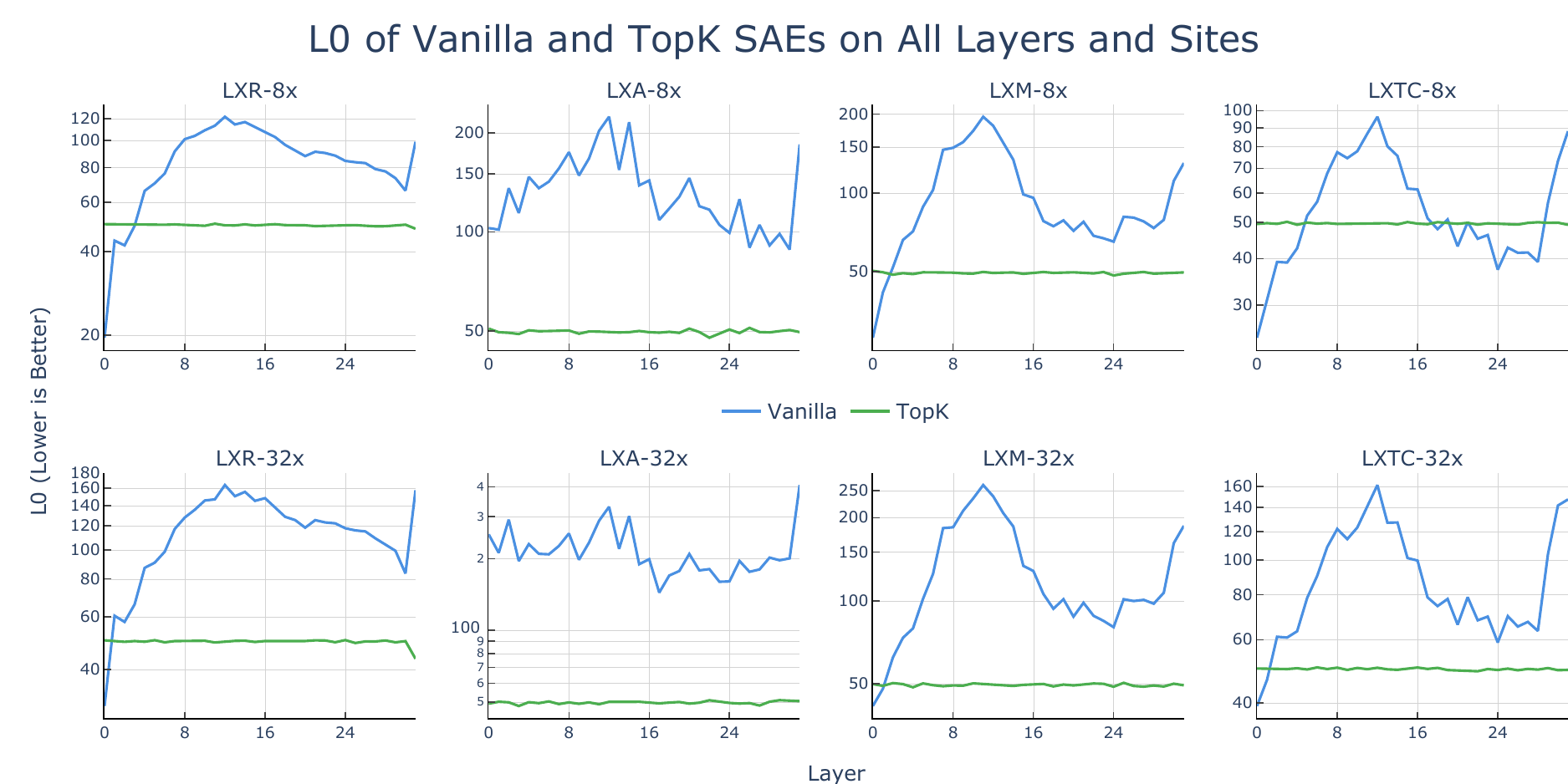}
       \label{fig:L0}
   \end{subfigure}

   \begin{subfigure}{\textwidth}
       \centering
       \includegraphics[width=\linewidth]{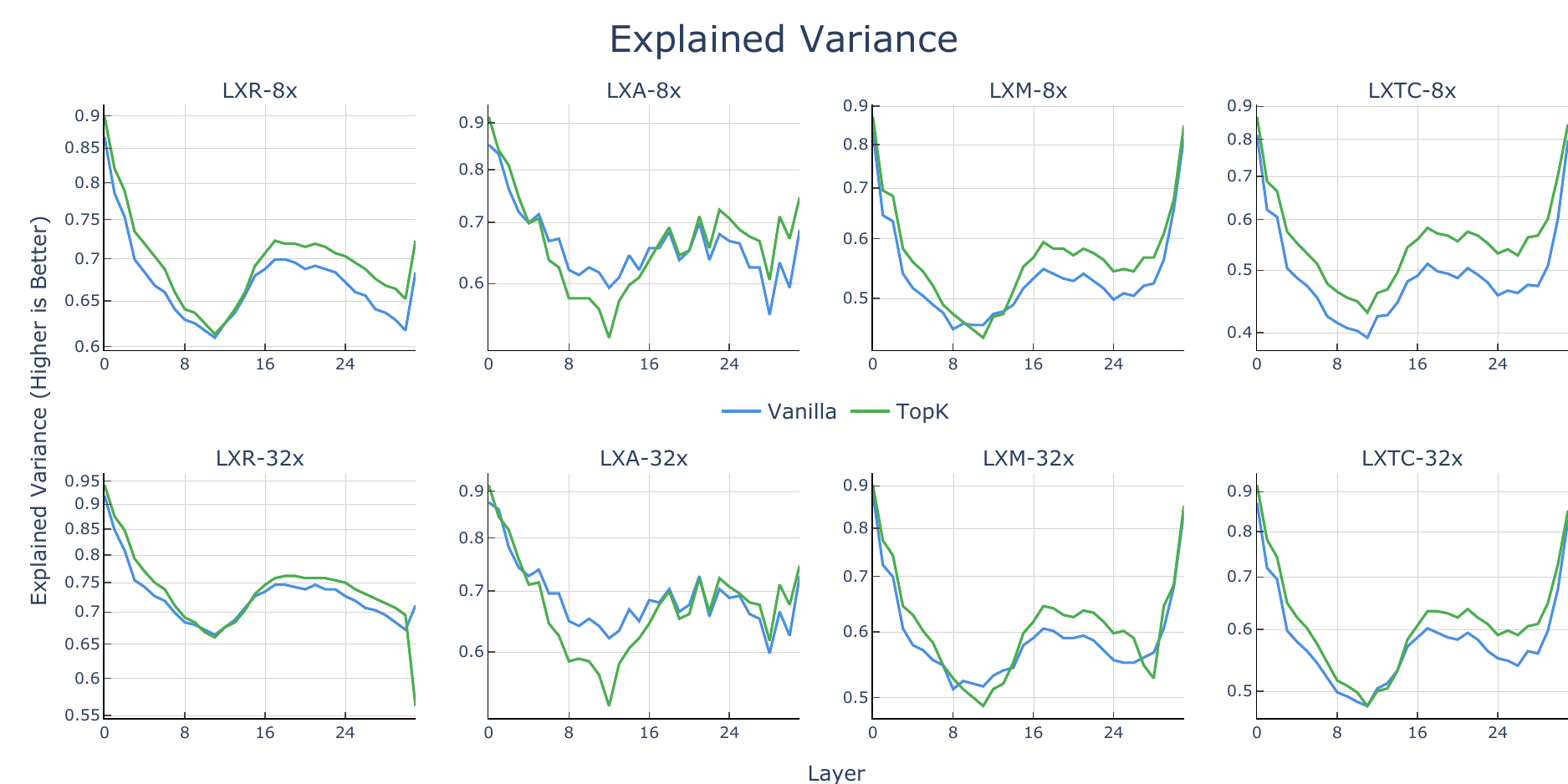}
       \label{fig:explained_variance}
   \end{subfigure}

   \begin{subfigure}{\textwidth}
       \centering
       \includegraphics[width=\linewidth]{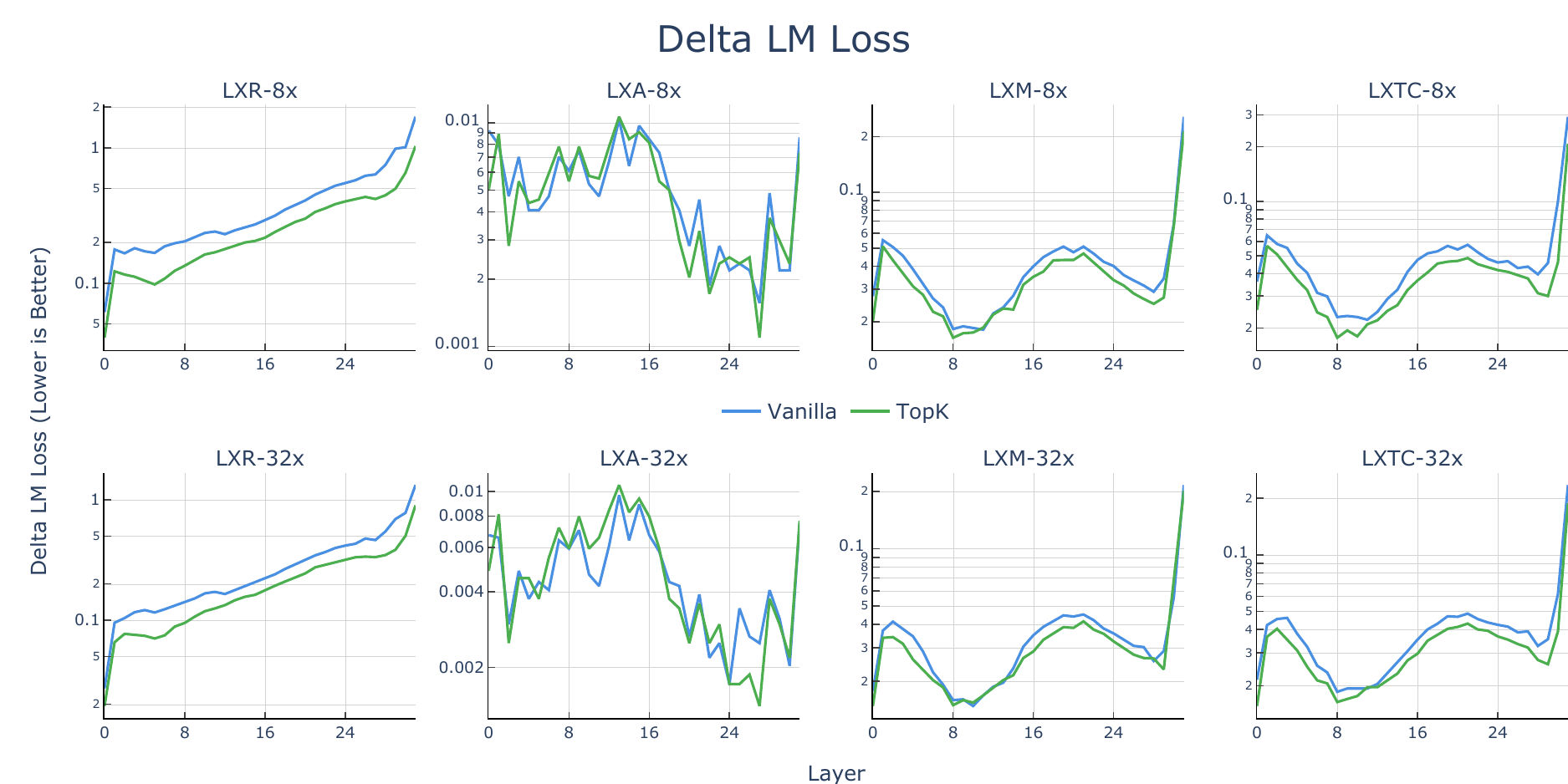}
       \label{fig:delta_ce_loss}
   \end{subfigure}

   \caption{All 256 SAEs are evaluated on L0 sparsity (upper), 
   explained variance (middle) and Delta LM loss (lower).}
   \label{fig:overall eval}
\end{figure}

\section{Ablation Studies}

\paragraph{Post-Processing Does Not Hurt Performance.}

\begin{figure}[h]
   \centering
   \includegraphics[width=\textwidth]{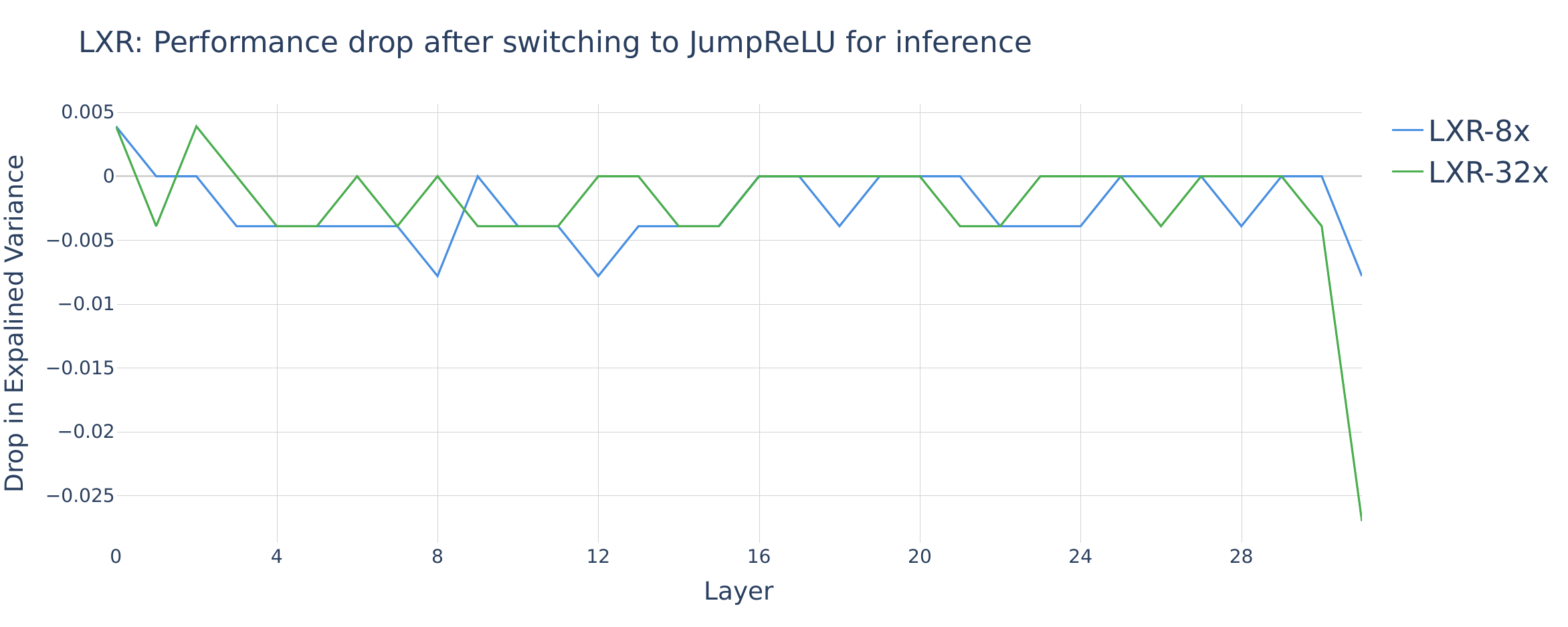}
   \caption{Drop in performance is negligible after switching to JumpReLU inference mode.}\label{fig:jumprelu}
\end{figure}

We find that the post-processing step of Section~\ref{sec:post-training} does not hurt sparsity-fidelity
trade-off. All post training transformations are computationally equivalent except for JumpReLU inference mode,
where we do not use TopK sparsity constraint but instead find a threshold where K features fire at each position
\emph{in expectation}. We evaluate the L0 sparsity and MSE efficiency of all 32 LXR-8x-TopK SAEs with and without
JumpReLU inference mode. The results are nearly identical, as shown in Figure~\ref{fig:jumprelu}.

\begin{figure}[h]
   \centering
   \begin{subfigure}[b]{0.48\textwidth}
      \centering
      \includegraphics[width=\linewidth]{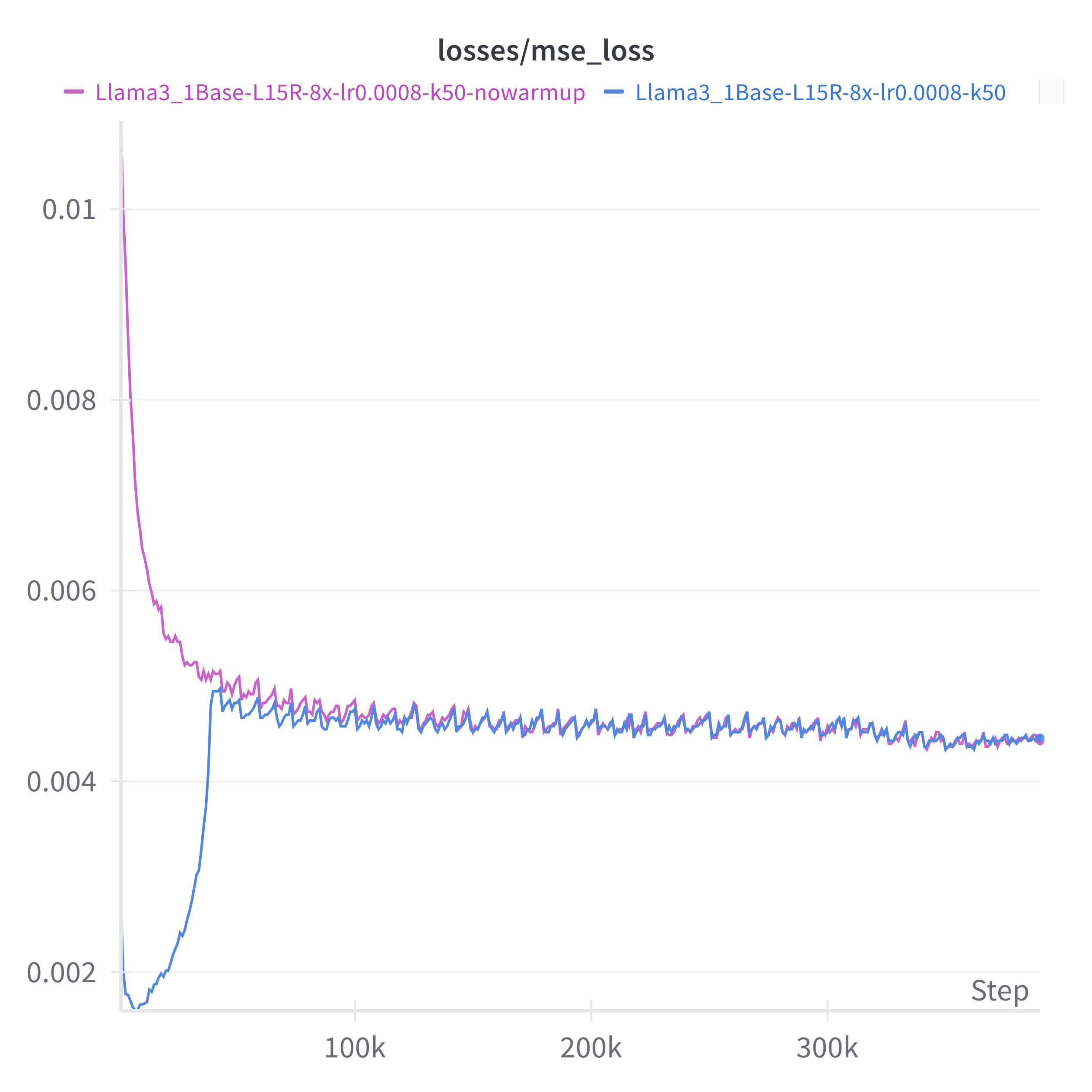}
   \end{subfigure}
   \vspace{0.2cm}
   \begin{subfigure}[b]{0.48\textwidth}
      \centering
      \includegraphics[width=\linewidth]{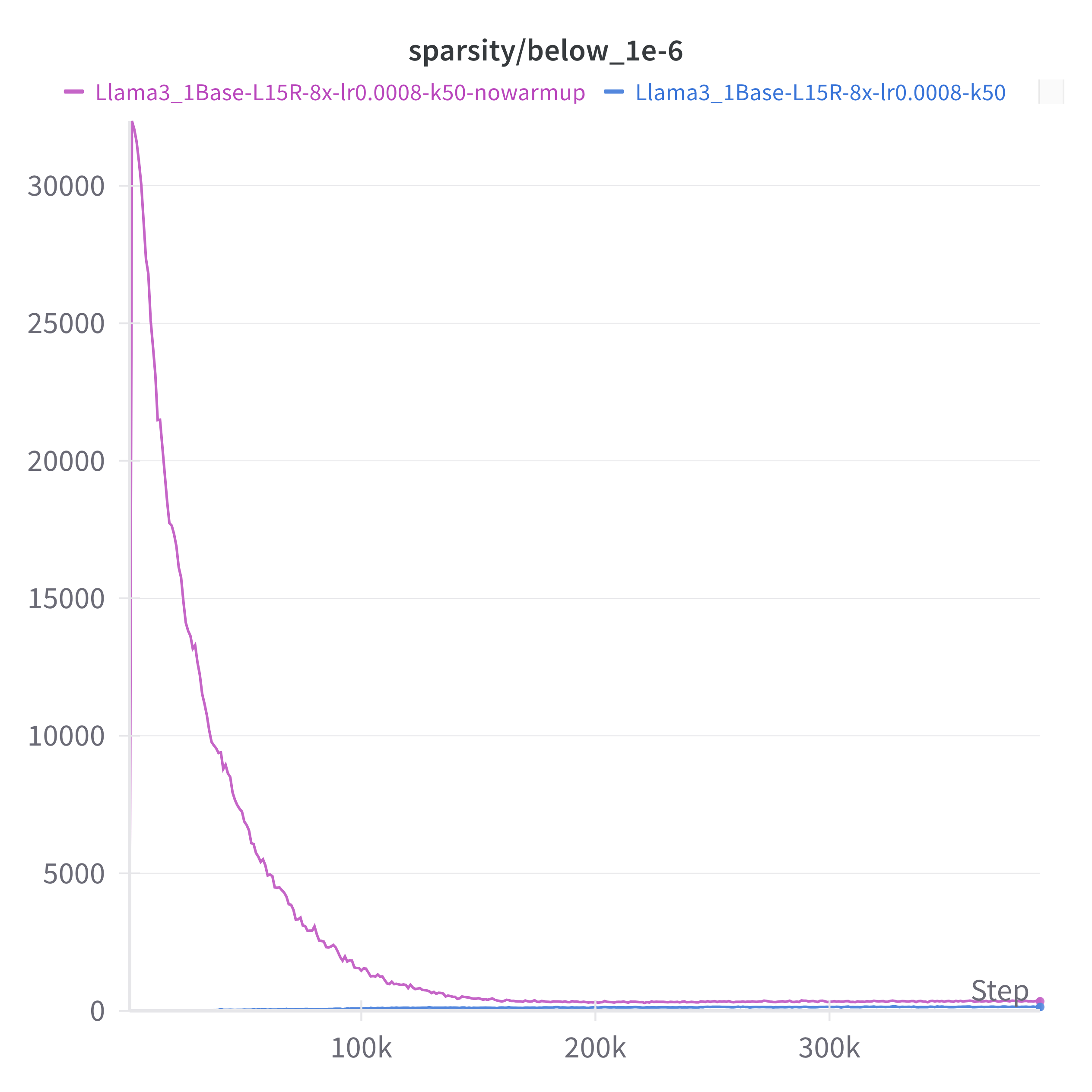}
   \end{subfigure}
   \caption{Convergence of L15R-8x-TopK SAEs with and without K-Annealing.}\label{fig:warmup_k}
   
\end{figure}

\paragraph{K-Annealing Accelerates Convergence.}
Figure~\ref{fig:warmup_k} shows the training curve of L15R-8x-TopK SAEs with and without annealing K at start of training.
By annealing K from 4096 to 50 in the first 10\% of the training steps, we find that the SAEs latents are active
during the whole training process (almost all features fire at least once over 1e6 tokens). In comparison,
the SAE features without K-annealing do not begin to activate until about 1/3 of the training steps.
Besides, both training curves converge to the same MSE loss but the K-annealing curve converges faster.

\paragraph{Including Decoder L2 Norm Improves Reconstruction?}
We also ablated the inclusion of decoder L2 norm in the topk sparsity constraint
(Equation~\ref{eq:encoder-our-topk}) with all other hyperparameters fixed. However,
the original version reported by \citet{gao2024oaisae} where the decoder L2 norm is fixed
to 1 after each training step is significantly worse in terms of MSE loss
(at least a 0.05 drop in variance explained). We suspect there are bugs in replicating
this baseline and leave it as future work.

\section{Prompt Design for Feature Interpretability}

\begin{Verbatim}[fontsize=\small, label=GPT-4o Prompt]
We are analyzing the activation levels of features in a neural network,
where each feature activates certain tokens in a text. Each token's
activation value indicates its relevance to the feature, with higher
values showing stronger association. Your task is to give this feature
a  monosemanticity score based on the following scoring rubric:

Activation Consistency

5: Clear pattern with no deviating examples

4: Clear pattern with one or two deviating examples

3: Clear overall pattern but quite a few examples not fitting that
pattern

2: Broad consistent theme but lacking structure

1: No discernible pattern

Consider the following activations for a feature in the neural
network. Activation values are non-negative, with higher values
indicating a stronger connection between the token and the
feature. You only need to return a number. It
represents your score for feature monosemanticity.

[Context]
Sentence 1: 
<START>
 students	0.0
 to	0.0
 build	0.0
 and	0.0
 repair	0.0
 all	0.0
 varieties	0.0
 of	0.0
 aviation	0.0
 instrumentation	0.0
 .	0.0
 The	0.0
 se	0.0
 ap	0.0
 lane	0.0
 base	0.0
 continued	0.0
 operations	0.0
 during	0.0
 the	0.0
 war	3.1
 as	0.0
 an	0.0
 all	0.0
 -f	0.0
<END>

...
\end{Verbatim}

\end{document}